\pdfoutput=1
\documentclass[preprint,12pt]{elsarticle}

\usepackage{graphicx}
\graphicspath{{./}}
\usepackage{xcolor}
\usepackage{algorithm}
\usepackage{algpseudocode}
\usepackage{booktabs}
\usepackage{adjustbox}
\usepackage{amssymb}
\usepackage{amsmath}
\usepackage{url}
\usepackage{hyperref}
\usepackage{cleveref}
\usepackage{float}
\usepackage{lineno}
\usepackage{etoolbox}
\usepackage{tikz}
\usepackage{subfigure}
\usepackage[normalem]{ulem}

\AtBeginEnvironment{document}{}
\newcommand{\startmainlinenumbers}{}
\newcommand{\stopmainlinenumbers}{}

\journal{Advanced Engineering Informatics}

\makeatletter
\def\ps@pprintTitle{%
  \let\@oddhead\@empty
  \let\@evenhead\@empty
  \def\@oddfoot{\parbox[b]{\textwidth}{\footnotesize\raggedright
    This is the accepted manuscript of an article published in Advanced Engineering Informatics, vol.~77, 105394 (2027).
    The final version is available at \url{https://doi.org/10.1016/j.aei.2026.105394}.
    \copyright~2026. This manuscript version is made available under the CC-BY-NC-ND 4.0 license
    \url{https://creativecommons.org/licenses/by-nc-nd/4.0/}.}}%
  \let\@evenfoot\@oddfoot}
\makeatother

\begin{document}

\begin{frontmatter}

\title{Resolving Primitive-Sharing Ambiguity in Long-Tailed TLS-Based Industrial MEP Point Cloud Segmentation via Spatial Context Constraints}

\author[label1]{Chao Yin}
\author[label2]{Qing Han}
\author[label3]{Zhiwei Hou}
\author[label1]{Yue Liu}
\author[label1]{Anjin Dai}
\author[label1,label3]{Hongda Hu}
\author[label1,label4]{Ji Yang\corref{cor1}}
\author[label5,label6]{Wei Yao\corref{cor2}}

\cortext[cor1]{Corresponding author. Email: yangji@gdas.ac.cn}
\cortext[cor2]{Corresponding author. Email: wyao@iue.ac.cn}

\affiliation[label1]{organization={Guangzhou Institute of Geography, Guangdong Academy of Sciences}, city={Guangzhou}, country={China}}
\affiliation[label2]{organization={School of Geography and Tourism, Hengyang Normal University}, city={Hengyang}, country={China}}
\affiliation[label3]{organization={Southern Marine Science and Engineering Guangdong Laboratory (Guangzhou)}, city={Guangzhou}, country={China}}
\affiliation[label4]{organization={Guangzhou iMapCloud Intelligent Technology Co., Ltd.}, city={Guangzhou}, country={China}}
\affiliation[label5]{organization={State Key Laboratory of Regional and Urban Ecology, Institute of Urban Environment, Chinese Academy of Sciences}, city={Xiamen}, country={China}}
\affiliation[label6]{organization={School of Engineering and Design, Technical University of Munich}, city={Munich}, country={Germany}}

\begin{abstract}
In terrestrial laser scanning (TLS)-based mechanical, electrical, and plumbing (MEP) point cloud segmentation, safety-critical components such as reducers and valves are persistently misclassified, blocking reliable engineering knowledge extraction. This stems from a dual crisis—extreme class imbalance (215:1) compounded by geometric ambiguity, since most tail classes share cylindrical primitives with dominant head classes—that existing frequency-based re-weighting methods cannot resolve.
We propose spatial context constraints that exploit neighborhood prediction consistency to disambiguate locally similar structures. Our approach extends Class-Balanced (CB) Loss with two architecture-agnostic mechanisms: Boundary-CB, an entropy-based constraint that emphasizes ambiguous boundaries and encodes an MEP assembly-topology prior, and Density-CB, a density-based constraint that compensates for scan-dependent variations and encodes TLS sensor-physics knowledge. Both operate at the loss level and require no backbone modifications; we evaluate them with ResPointNet++.
On the Industrial3D dataset (612.7M labelled points from water treatment facilities), our method achieves 55.74\% mIoU on the evaluation split, which also served as the validation split, exceeding the strongest of three backbone baselines (39.83--52.48\% mIoU), with a 21.7\% relative improvement on tail-class performance (29.59\% vs.\ 24.32\%) while preserving head-class accuracy (88.14\%). Reducer improves from 0\% to 21.12\% IoU and valve by 24.3\% relative. Evaluated per area, the gain comes from the area rich in these primitive-sharing components (tail mIoU 21.4\% to 26.5\%), while the other area's mIoU is 0.6 points lower. Spatial context constraints thus reduce primitive-sharing errors in the target MEP setting, supporting more reliable identification of safety-critical components for Digital Twin and Scan-to-BIM applications. Code: https://github.com/PointCloudYC/LongTail3D.git.
\end{abstract}

\begin{highlights}

\item Defines a dual crisis in TLS-based MEP scenes: 215:1 imbalance, geometric ambiguity.
\item Proposes engineering knowledge-informed Boundary-CB and Density-CB constraints.
\item Interprets both constraints via intuitive metaphors and mathematical visualization.
\item Achieves 55.74\% mIoU on Industrial3D dataset with 21.7\% tail-classes improvement.
\item Improves reducer IoU from 0\% to 21.12\%; valve improves by 24.3\% relative.
\end{highlights}

\begin{keyword}
Spatial context constraints \sep Primitive-sharing ambiguity \sep Long-tailed point cloud segmentation \sep TLS-based MEP scenes \sep Class-balanced loss \sep Neighborhood entropy \sep Digital twin
\end{keyword}

\end{frontmatter}

\startmainlinenumbers
\section{Introduction}
Digital Twin construction for the mechanical, electrical, and plumbing (MEP) systems of industrial facilities requires accurate 3D point cloud semantic segmentation (PCSS) to extract functional component information~\cite{mirzaei2022pcml,guo2020}. However, current methods systematically fail on safety-critical components like valves and reducers. These failures occur for two reasons: such components are rare in training data, and they share identical local geometry with dominant structures like pipes. We identify a  specific challenge in  terrestrial laser scanning (TLS)-based industrial MEP point clouds: a ``dual crisis'' where extreme class imbalance (215:1 ratio) compounds geometric ambiguity from primitive sharing between head and tail classes.

This dual crisis affects practical engineering applications. Digital Twin construction, facility management, and decision support systems rely on automated knowledge extraction~\cite{ma2018reconstruction,chen2023improving,shao2024urban}~\cite{xu2025interception,ge2025resilient}. While 3D acquisition technologies (LiDAR, photogrammetry) capture geometric data well, extracting semantic knowledge—identifying functional components like valves and pumps—remains difficult~\cite{agapaki2020,yin2021,yin2022}~\cite{jing2024mep}~\cite{yue2025enhancing}~\cite{cheng2025coating}. Point Cloud Semantic Segmentation (PCSS) converts raw geometric data into structured knowledge for predictive maintenance and safety monitoring~\cite{yue2024deep,wang2019automatic,wang2022vision}. When PCSS fails on safety-critical components, the entire knowledge extraction pipeline breaks down.

The proposed constraints encode engineering knowledge directly into this learning process. Boundary-CB reflects an MEP assembly-topology prior: functional component boundaries often coincide with geometric primitive transitions, such as pipe-to-reducer and pipe-to-valve interfaces. By emphasizing prediction-uncertain boundary regions, the model learns spatial co-occurrence patterns that define engineering assembly topology rather than relying on class frequency alone. Density-CB reflects TLS acquisition physics: point density varies with scanner range, incidence angle, and occlusion, so compensating for residual density variation after grid subsampling injects knowledge of the sensing process into loss modulation.

\begin{figure*}[!htbp]
    \centering
    \includegraphics[width=0.95\linewidth]{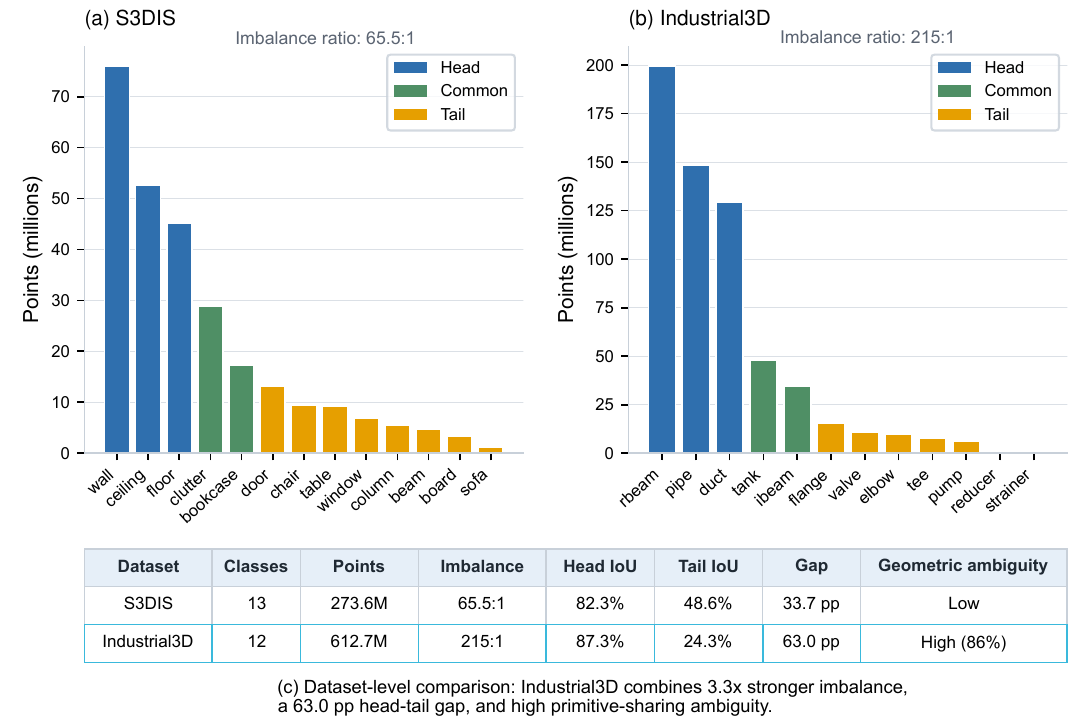}
    \caption{Dataset comparison characterizing Industrial3D's dual crisis. Panels (a-b) show class distributions for S3DIS (indoor) and Industrial3D (water-treatment MEP), with head (blue), common (green), and tail (orange) classes sorted by frequency. Panel (c) provides a quantitative comparison table: (1) \textit{Statistical severity:} 3.3$\times$ greater point-count imbalance than S3DIS (215:1 vs. 65.5:1 by unique labelled point count); (2) \textit{Performance impact:} a 63.0 percentage-point head--tail gap (87.3\% vs. 24.3\% tail IoU), compared with 33.7 pp in S3DIS; and (3) \textit{Geometric ambiguity:} six of seven tail classes (86\%) share cylindrical primitives with Pipe. These measurements define the dual crisis studied in this TLS-based MEP benchmark. Abbreviation ``rbeam'' denotes rectangular beam.}
    \label{fig:longtail_comparison}
\end{figure*}

Modern architectures perform well on benchmarks with moderate class imbalance, such as S3DIS~\cite{armeni2016} ( approximately 65:1 by unique labelled point count) and ScanNet~\cite{dai2017} (20--30:1 ratios). These include PointNet/PointNet++~\cite{qi2017pointnet,qi2017pointnetpp}, transformers~\cite{wu2024point}, and lightweight designs like RandLA-Net~\cite{hu2020randlanet}. However, the water-treatment MEP scenes studied here present a different challenge: extreme class imbalance (>200:1) combined with geometric ambiguity where tail classes share primitives with head classes. \autoref{fig:longtail_comparison} compares S3DIS and Industrial3D datasets directly. Industrial3D shows about  3.3$\times$ greater point-count imbalance and significantly higher geometric ambiguity. Standard models fail on safety-critical components in this benchmark.

Frequency-based re-weighting methods address extreme imbalance in 2D vision. Class-Balanced (CB) Loss~\cite{cui2019class} introduces the ``effective number of samples'' to formalize how additional samples provide diminishing marginal information. For a class $c$ with $n_c$ samples, the effective number is $E_c = (1-\beta^{n_c})/(1-\beta)$, where $\beta$ controls information overlap. This formulation creates a ceiling effect: for large $n_c$, $E_c$ plateaus. This prevents the extreme gradients of naive inverse-frequency schemes while still boosting rare classes.

This 2D solution raises a question for TLS-based MEP point clouds: what factors beyond point count affect a class's effective representation? CB Loss treats all samples uniformly. This assumption fails when geometrically distinct objects (e.g., a Pipe vs. a Reducer) share identical local primitives (cylinders). Our analysis of the Industrial3D dataset reveals a structural pattern: head classes (beams, pipes; 77\% of points) are predominantly long, continuous structures, whereas tail classes (each below 3\% of points) exhibit geometric heterogeneity. Tail classes split into two categories. \emph{Primitive-similarity} classes (e.g., elbows, reducers) are locally indistinguishable from head-class pipes. \emph{Composite-tail} classes (e.g., pumps, valves) are multi-part components whose individual parts share geometry with head-class primitives.

To resolve this geometric ambiguity, we introduce spatial context constraints that use neighborhood prediction consistency to disambiguate locally similar structures. We formalize two complementary constraints extending the Class-Balanced (CB) framework: the Boundary-CB constraint adaptively emphasizes ambiguous boundaries by measuring neighborhood entropy, while the Density-CB constraint normalizes for scan-dependent density variations. Both constraints operate at the loss level and can be added to other point cloud pipelines without network modifications.

Our contributions are:
\begin{itemize}
    \item Problem formalization: We formally characterize the ``dual crisis'' in TLS-based water-treatment MEP point clouds—extreme statistical imbalance (215:1) compounded by geometric ambiguity where most of tail classes share primitives with head classes—and identify structural heterogeneity among tail classes (composite-tail vs.\ primitive-similarity).
    \item Spatial context constraints: We introduce Boundary-CB, which leverages neighborhood prediction consistency to resolve ambiguity from shared primitives, and Density-CB, which compensates for scan-dependent density variations, establishing geometry-aware long-tailed 3D learning.
    \item Interpretability analysis: Deep learning models are often criticized as black boxes that obscure their decision-making processes. We provide comprehensive interpretability through two complementary visualizations—intuitive metaphors (resource allocation, neighborhood consensus) and rigorous mathematical foundations—making the proposed constraints transparent and accessible to both researchers and practitioners.
    \item Architecture-agnostic integration: Our constraints function as plug-and-play modules that require no architectural modifications (evaluated here with ResPointNet++), enabling integration into existing MEP segmentation pipelines.
    \item Empirical validation: On the predefined Industrial3D Area~6+12 evaluation split, Boundary-CB achieves 55.74\% mIoU and 29.59\% tail-class mIoU (+21.7\% relative), improving recognition of safety-critical components while preserving head-class accuracy—avoiding the typical head-tail trade-off.
\end{itemize}

This paper is organized as follows: \autoref{sec:related_work} reviews related work, \autoref{sec:methodology} details our proposed spatial context constraints, \autoref{sec:experiments} presents experimental validation, \autoref{sec:discussion} discusses implications for engineering informatics, and \autoref{sec:conclusion} concludes.

\section{Related Work}
\label{sec:related_work}
We review point cloud semantic segmentation and long-tailed learning methods to explain why existing approaches fail to address geometric ambiguity from primitive sharing in MEP point clouds.

\subsection{Fully Supervised Point Cloud Semantic Segmentation}
\label{sec:fully_supervised}
Fully supervised methods learn from completely annotated data and dominate 3D semantic segmentation. These approaches fall into three categories: point-based, projection-based, or voxel-based~\cite{guo2020}. Point-based networks operate directly on raw point clouds and have become the standard due to their architectural simplicity, efficiency, and ability to preserve geometric fidelity.

PointNet~\cite{qi2017pointnet} pioneered permutation-invariant architectures for point sets. PointNet++~\cite{qi2017pointnetpp} introduced hierarchical feature learning to capture multi-scale local geometry, a concept that underpins most modern convolutional neural network (CNN) architectures. Subsequent work produced more powerful local aggregation operators: RandLA-Net~\cite{hu2020randlanet} uses random sampling and attention mechanisms for efficiency, KPConv~\cite{thomas2019} employs specialized kernel-based convolutions, and ResPointNet++~\cite{yin2021} integrates residual connections for complex industrial MEP scenes. These methods work well on balanced benchmarks but can degrade under the severe class imbalance found in MEP datasets such as Industrial3D.

Recent work has improved accuracy and efficiency: sparse voxel backbones using submanifold sparse convolutions~\cite{graham2017}, transformer-based designs such as Point Transformer v3~\cite{wu2024point}, unified segmentation frameworks like OneFormer3D~\cite{kolodiazhnyi2024oneformer3d}, boundary-focused feature modeling (BFANet)~\cite{zhao2025bfanet}, and analyses of local aggregation operators~\cite{liu2020}. Large-scale annotation tools like ARKit LabelMaker~\cite{ji2025arkit} also expand dataset scale and supervision breadth. However, these advances primarily enhance representation capacity. They do not explicitly resolve long-tailed statistical bias or geometric ambiguity from primitive sharing in MEP point clouds.

\subsection{Deep Long-Tailed Learning}
\label{sec:long_tailed}
Long-tailed learning addresses training on datasets with severe class imbalance, where a few ``head'' classes dominate the distribution and numerous ``tail'' classes are under-represented~\cite{zhang2023longtail}. Core strategies include re-sampling, data augmentation, and loss re-weighting~\cite{more2016,zhang2017mixup,verma2019,cui2019class}.

Following the survey taxonomy of long-tailed learning~\cite{zhang2023longtail}, these strategies can be grouped into data-level re-sampling, loss-level re-weighting, feature-level augmentation or transfer, and contrastive or generative synthesis. We focus on loss-level re-weighting because it preserves the dense point-wise supervision setting while allowing the loss to encode spatial information that class frequency alone ignores.

Re-weighting vs.\ re-sampling. In 2D vision, re-sampling (over-/under-sampling) can reduce bias but often overfits minority samples or discards informative majority examples. For point clouds, these problems get worse: point- or patch-level over-sampling distorts density statistics and breaks geometric continuity; under-sampling disrupts the geometric continuity required by kNN/ball-query operators. We therefore prefer loss re-weighting, which preserves native geometry while correcting gradient contributions.

This choice is particularly important for TLS-based industrial MEP segmentation. Generative synthesis approaches such as RealNet~\cite{zhang2024realnet} and unified anomaly-detection paradigms such as AnomalyMoE~\cite{gu2026anomalymoe} would require enough tail-class examples to model not only the local 3D geometry of reducers, strainers, and valves, but also their physically valid scene context, attachment direction, and orientation relative to pipes. Under a 215:1 imbalance, those conditions are not reliably met, and the anomaly-detection setting itself targets normal-vs-deviation scoring rather than dense multi-class labelling. Few-shot episode construction is also mismatched with dense per-point scene segmentation, where support and query sets do not naturally separate at object level inside large registered facilities. Sim-to-real transfer~\cite{wu2023sim2real} offers a complementary route via simulated industrial scenes, but its domain-gap and label-noise challenges remain orthogonal to the loss-level decision studied here. Contrastive pre-training can complement the backbone, but it does not replace the supervised loss term that must decide how much each ambiguous point contributes during training.

Class-Balanced (CB) Loss~\cite{cui2019class} formalizes re-weighting via the ``effective number of samples,'' stabilizing training compared to naive inverse frequency scaling or smooth variants like inverse square root scaling~\cite{rosu2020,cortinhal2020,hu2022sensaturban}. Focal Loss~\cite{lin2017} complements CB by down-weighting easy instances at the instance level. Their combination (CB+Focal) forms our strong frequency-based baseline.

Augmentation. MixUp/CutMix-style augmentation~\cite{zhang2017mixup,verma2019} and standard 3D transforms (rotation, scaling, jitter) improve generalization but do not directly correct class imbalance. They complement re-weighting approaches. However, frequency-only methods treat class members uniformly and ignore geometric factors—a limitation that becomes critical in MEP point clouds, where classes share geometric primitives.

\subsection{Long-Tailed Learning in 3D Point Clouds}
Long-tailed learning for 3D point clouds is an emerging research area. Challenges exceed those in 2D due to unstructured representation, variable density, and occlusion—conditions that are pronounced in the evaluated TLS-based MEP environments, with severe class imbalance (>200:1), complex structures, and rare but critical components (e.g., pumps, valves).

Recent 3D approaches. Contemporary methods mitigate imbalance through sampling and optimization designs: two-stage/target-guided strategies for rare classes~\cite{zhang2022two,zhang2023target}, adaptive weight constraints with class-aware sampling~\cite{lahoud2024}, class-imbalanced semi-supervised schemes with decoupled optimization~\cite{li2024decoupling}, and target-aware attentional networks for rare class segmentation~\cite{zhang2025target}. These techniques primarily adjust data distribution or training objectives under label scarcity. They generally do not explicitly model geometric ambiguity and are complementary to our geometry-aware re-weighting.

The geometric dimension of imbalance. Beyond statistical frequency, 3D point clouds introduce geometric ambiguity from systematic primitive-sharing between classes. Our analysis reveals a structural pattern: head classes (beams, pipes, ducts) have long-shaped, continuous geometry, whereas tail classes split into two categories. First, \emph{composite-tail} classes (pumps, valves) are complex assemblies whose components share primitives with head classes, creating confusion at part boundaries. Second, \emph{primitive-similarity tail} classes (elbows, tees, reducers) are simple objects locally indistinguishable from head-class pipes. This creates challenges that frequency-based methods cannot address:
\begin{enumerate}
    \item Compositional ambiguity: Composite-tail objects have boundary regions where local geometry matches head-class geometry, creating confusion at multi-part interfaces.
    \item Intrinsic similarity: Primitive-similarity tail objects (e.g., cylindrical elbows) share fundamental geometric primitives with head-class pipes. Local features are insufficient for discrimination, requiring global context.
\end{enumerate}

Limitations of existing approaches. Recent 3D long-tailed learning methods~\cite{zhu2019,zhang2022two,zhang2023target,lahoud2024,li2024decoupling} and modern point cloud architectures~\cite{qi2017pointnet,thomas2019,huang2020deep,wang2025test} typically adapt 2D frequency-based techniques without accounting for geometric factors. This addresses the \textit{symptom} (statistical imbalance) but overlooks the \textit{root cause} of misclassification in 3D data—geometric ambiguity from shared primitives. Our work explicitly leverages spatial context constraints to resolve this geometric ambiguity. We introduce constraints that measure neighborhood prediction consistency to identify and emphasize ambiguous boundaries, addressing both compositional and primitive-similarity ambiguity. Boundary-sensitive segmentation~\cite{zhao2025bfanet} and unified 3D frameworks~\cite{kolodiazhnyi2024oneformer3d} improve representations but do not mitigate frequency-driven bias that overwhelms tail classes. Our spatial context constraints are orthogonal and complementary to such architectural advances.

\subsection{Industrial MEP Point Cloud Analysis}
\label{sec:industrial_pc}
TLS-based MEP point cloud segmentation presents domain-specific challenges. Our comparison of six benchmarks (see Supplementary Material) shows that Industrial3D combines extreme statistical imbalance with severe primitive-sharing ambiguity. Indoor architectural and outdoor urban benchmarks differ in taxonomy, acquisition, and spatial structure, so the comparison characterizes the target setting rather than asserting that every industrial dataset has the same properties. In Industrial3D, six of seven tail classes (86\%) share cylindrical geometry with dominant head classes, creating local indistinguishability. This geometric homogeneity, combined with a 215:1 imbalance ratio, defines the dual crisis evaluated here. Complementary work in AEC/MEP scenarios—vision-assisted and vision-foundation-model-based BIM reconstruction~\cite{wang2022vision,wang2024omni}, hierarchical segmentation with vision-language reasoning~\cite{li2025integrating}, robot-assisted mobile scanning~\cite{hu2023robot}, BIM-to-scan domain adaptation~\cite{hu2024automated}, label-efficient training~\cite{yin2023,wang2023one,wang2022new}, and cross-modal 2D--3D fusion~\cite{yue20252d} or BIM-generated synthetic supervision~\cite{yue2025enhancing}—primarily reduces annotation cost or domain gaps. Recent AEI studies further show the engineering relevance of point-cloud segmentation for spraying trajectory planning~\cite{cheng2025coating}, BIM-generated supervision for MEP semantic segmentation~\cite{yue2025enhancing}, and digital-twin reasoning for infrastructure operation~\cite{xu2025interception,ge2025resilient}. These approaches do not directly correct the long-tailed statistical and geometric imbalance that we target.

The closest recent MEP semantic-segmentation dataset by Jing et al.~\cite{jing2024mep} contains 92.1M labelled points over 9 MEP classes from commercial and residential buildings. Industrial3D differs by targeting operational water-treatment facilities, using 612.7M labelled TLS points, and emphasizing facility-scale primitive sharing among pipes, reducers, valves, pumps, beams, and ducts.

Even fine-grained taxonomies like ScanNet200~\cite{rozenberszki2022} (200 classes, 145:1 imbalance) exhibit only 32\% geometric ambiguity and benefit from RGB texture, which is less discriminative for the metallic components of MEP scenes~\cite{yin2026industrial3d}. MEP applications require reliable segmentation of critical components for digital twins~\cite{yin2023,pierdicca2020}, predictive maintenance, and safety. Misclassifying a valve as a pipe can lead to incorrect flow analysis, faulty maintenance, or safety hazards. Existing long-tailed methods, developed for less imbalanced datasets, fail to address the geometric ambiguity inherent in MEP systems, where head-class structures (pipes, beams) dominate both statistically (77\% of data) and geometrically (through shared primitives). Our work targets this dual crisis by introducing spatial context constraints that leverage neighborhood prediction consistency to resolve geometric ambiguity.

\section{Methodology}
\label{sec:methodology}
\begin{figure*}[htbp]
    \centering
    \includegraphics[width=0.95\linewidth]{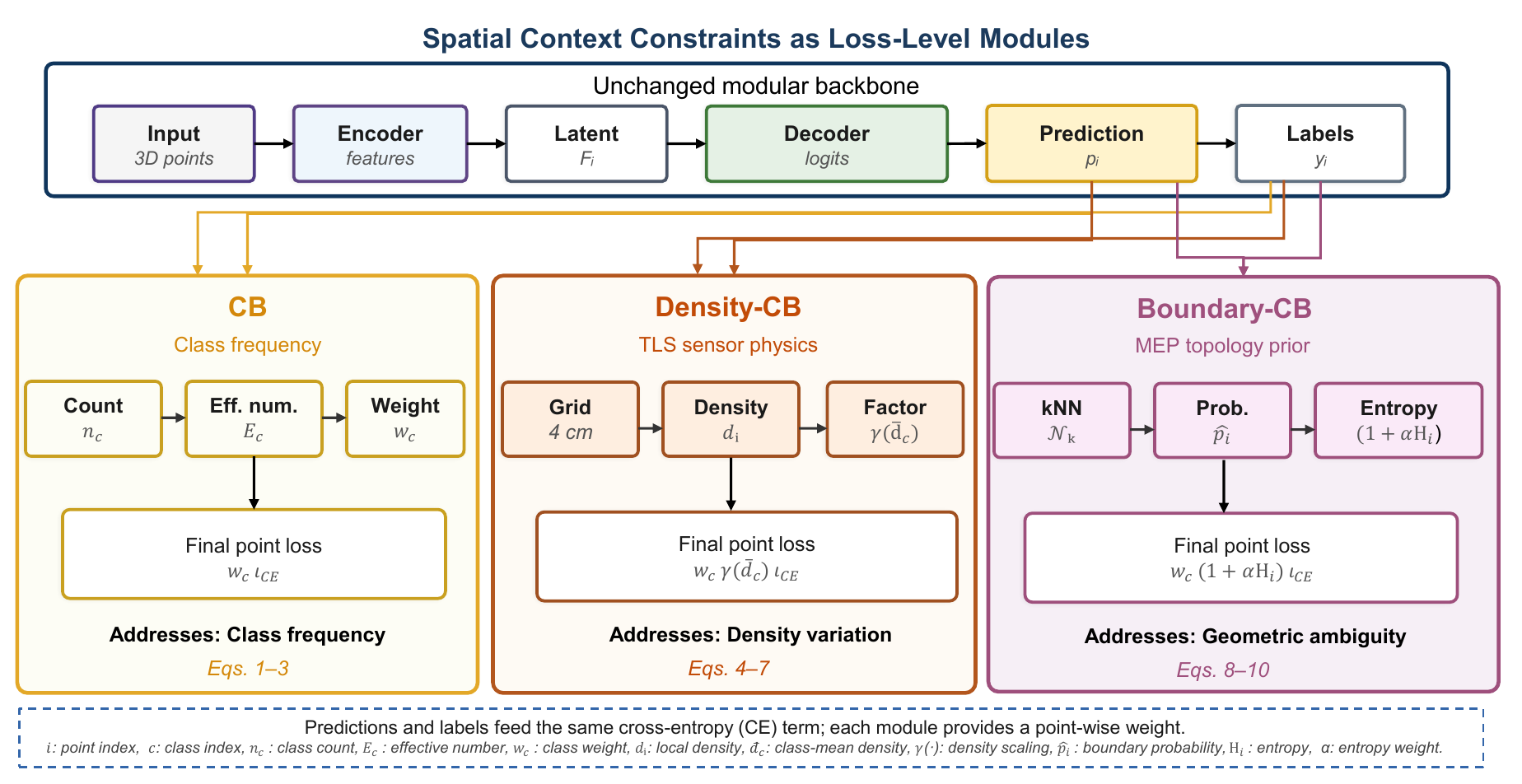}
    \caption{Spatial context constraints as loss-level modules. The top row shows the unchanged modular segmentation backbone: input points are encoded, decoded, and converted into per-point predictions, while labels provide the supervised target. The bottom row shows three parallel loss-weighting paths that all multiply the same cross-entropy term. CB uses class counts to compute the frequency weight~$w_c$ (Eqs.~\ref{eq:cb_effective}--\ref{eq:cb_loss}); Density-CB, motivated by TLS sensor-physics knowledge, adds local grid-density statistics to form~$\gamma(\bar{d}_c)$ (Eqs.~\ref{eq:local_density}--\ref{eq:ld_cb_loss}); and Boundary-CB, motivated by MEP assembly-topology priors, uses $k$NN neighborhood predictions to compute entropy~$H_i$ and the modulator~$(1+\alpha H_i)$ (Eqs.~\ref{eq:neighborhood_prob}--\ref{eq:bc_cb_loss}). The selected loss backpropagates through the unchanged backbone, so the contribution lies in loss modulation rather than network redesign. For clarity, the figure omits the focal factor~$\phi_i$ that Density-CB and Boundary-CB share with CB+Focal (Eqs.~\ref{eq:ld_cb_loss} and~\ref{eq:bc_cb_loss}).}
    \label{fig:framework_overview}
\end{figure*}

This section presents our spatial context constraints for resolving geometric ambiguity in long-tailed TLS-based MEP point clouds. We propose two architecture-agnostic mechanisms that extend the Class-Balanced framework by incorporating spatial context: the Boundary-CB constraint for resolving geometric ambiguity through neighborhood entropy, and Density-CB for normalizing scan-dependent density variations. \autoref{fig:framework_overview} illustrates how these spatial context constraints integrate into standard point cloud segmentation pipelines as plug-and-play modules without requiring backbone modifications. We first formalize the dual crisis problem (\S\ref{sec:problem_formulation}), review the Class-Balanced (CB) Loss baseline (\S\ref{sec:cb_loss_baseline}), then introduce our two spatial context constraints: Density-CB (\S\ref{sec:ld_cb_loss}) and Boundary-CB (\S\ref{sec:bc_cb_loss}). Finally, we specify evaluation metrics (\S\ref{sec:evaluation_metrics}).

\subsection{Problem Formulation: The Dual Crisis}
\label{sec:problem_formulation}
Consider a training dataset \(\mathcal{D} = \{(P_i, Y_i)\}_{i=1}^{N_{\text{scene}}}\) of \(N_{\text{scene}}\) point cloud scenes. Each scene \(P_i = \{p_1^{(i)}, \ldots, p_{M_i}^{(i)}\}\) contains \(M_i\) points, where \(p_j^{(i)} \in \mathbb{R}^3\) denotes the 3D coordinates of the \(j\)-th point in scene \(i\). The corresponding labels are \(Y_i = \{y_1^{(i)}, \ldots, y_{M_i}^{(i)}\}\), where \(y_j^{(i)} \in \{1, \ldots, C\}\) is the semantic class of point \(j\) from \(C\) total classes. For notational simplicity, we flatten all points across scenes into a single set of \(N = \sum_{i=1}^{N_{\text{scene}}} M_i\) points, \(\{p_1, \ldots, p_N\}\), with labels \(\{y_1, \ldots, y_N\}\).

In the evaluated TLS-based MEP scenes, the class frequency distribution is long-tailed. Let \(n_c = |\{i : y_i = c\}|\) be the number of points in class \(c\). Since \(n_c\) varies by orders of magnitude, standard training is dominated by head classes, leading to poor recognition of rare but critical tail classes.

Beyond statistical imbalance, 3D point clouds introduce geometric ambiguity: systematic confusion between semantically distinct classes that share local geometric primitives. Unlike 2D images, where object categories typically have distinguishable local features, many objects in TLS-based MEP scenes are composed of similar geometric primitives, differing mainly in global structure or context.

Our analysis of the Industrial3D dataset shows a critical structural prior. Head classes (\texttt{RectangularBeam}, \texttt{Pipe}, \texttt{Duct}), comprising 77\% of the dataset, are uniformly long-shaped, continuous structures. In contrast, tail classes (each below 3\% of points) display geometric heterogeneity and partition into two categories:

\begin{enumerate}
    \item Composite-tail classes (\texttt{Flange}, \texttt{Valve}, \texttt{Pump}, \texttt{Strainer}): These comprise complex multi-part assemblies that are structurally more intricate than head classes. For example, a pump consists of multiple components including a cylindrical housing, mounting flanges, and connection ports. While these objects have compositional complexity, their individual constituent components—cylinders, disks, and connecting elements—are geometrically identical to the primitives that compose head classes.
    
    \item Primitive-similarity tail classes (\texttt{Elbow}, \texttt{Tee}, \texttt{Reducer}): These are geometrically simple objects that share nearly identical local geometric primitives with the head-class \texttt{Pipe}. An elbow is a curved cylinder; a reducer is a tapered cylinder. At local neighborhood scales, these objects are indistinguishable from straight pipes.
\end{enumerate}

This structural pattern shows two mechanisms of geometric ambiguity. For \emph{primitive-similarity} tail classes, the ambiguity is \emph{intrinsic}: local features are insufficient for discrimination, requiring global context. For \emph{composite-tail} classes, the ambiguity is \emph{compositional}: boundary regions between constituent parts create local confusion. In both cases, models trained on severely imbalanced data default to head-class interpretations due to overwhelming statistical bias, as further evidenced by the dataset-level analysis in the Supplementary Material.

Statistical bias and geometric ambiguity together constitute the dual crisis studied in Industrial3D. Tail classes suffer from both statistical rarity (fewer training examples) and geometric confusion (local appearance similar to dominant head classes). This geometric dimension, particularly the structural heterogeneity among tail classes, is a central challenge in the evaluated MEP point cloud setting.

\subsection{Class-Balanced Loss Baseline}
\label{sec:cb_loss_baseline}
Our approach builds on Class-Balanced (CB) Loss~\cite{cui2019class}, a method for addressing statistical bias. The core idea is \textit{effective number of samples}, based on the observation that the information gain from additional examples diminishes as class frequency increases. For a class \(c\) with \(n_c\) points, the effective number of samples is:
\begin{equation}
\label{eq:cb_effective}
E_c = \frac{1 - \beta^{n_c}}{1 - \beta}
\end{equation}
where \(\beta \in [0,1)\) is a hyperparameter. The class-specific loss weight is the inverse of the effective number:
\begin{equation}
\label{eq:cb_weight}
w_c = \frac{1}{E_c} = \frac{1-\beta}{1-\beta^{n_c}}
\end{equation}
Let \(f_\theta : \mathbb{R}^3 \to \mathbb{R}^C\) be a segmentation network. The CB loss applies class-specific weights to the cross-entropy loss:
\begin{equation}
\label{eq:cb_loss}
\mathcal{L}_{\text{CB}} = \frac{1}{N} \sum_{i=1}^{N} w_{y_i} \cdot \ell_{\text{CE}}(f_\theta(p_i), y_i)
\end{equation}
where \(\ell_{\text{CE}}(\hat{y}, y) = -\log(\hat{y}_y)\) is the standard cross-entropy loss.

\begin{figure}[!htbp]
    \centering
    \includegraphics[width=0.85\linewidth]{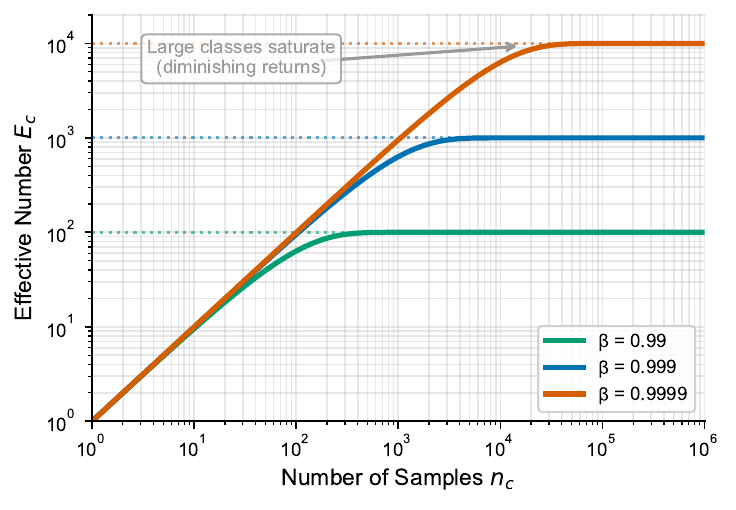}
    \caption{Ceiling effect of Class-Balanced Loss. The effective number of samples \(E_c\) saturates as actual sample count \(n_c\) increases, creating a ceiling effect that prevents extreme re-weighting. Different \(\beta\) values control the saturation rate: smaller \(\beta\) (e.g., 0.99) saturates quickly, while larger \(\beta\) (e.g., 0.9999) allows more gradual saturation. With \(\beta=0.9999\), classes beyond 10,000 samples receive nearly identical weights, stabilizing training while still emphasizing rare classes.}
    \label{fig:cb_ceiling_effect}
\end{figure}

As illustrated in ~\autoref{fig:cb_ceiling_effect}, CB Loss has an implicit ``ceiling effect.'' As \(n_c\) grows, \(E_c\) approaches \(1/(1-\beta)\). With \(\beta=0.9999\), this ceiling is 10,000, meaning a head class with 1,000,000 points and one with 10,000 points receive nearly identical weights. This design prevents extreme re-weighting that characterizes naive inverse-frequency schemes, while still providing gradient emphasis for genuinely rare tail classes.

Integration with Focal Loss. CB Loss is often combined with Focal Loss~\cite{lin2017}, which down-weights well-classified examples via the modulating factor \((1-\hat{p}_t)^{\gamma_{\mathrm{f}}}\). This provides complementary re-weighting at the class level (CB) and instance level (Focal). We compare against this strong CB+Focal baseline.

Limitations of frequency-only methods. Despite these advantages, frequency-only approaches overlook a critical dimension: the geometric structure of 3D data. They treat all points within a class uniformly, ignoring that a point from a rare \texttt{Pump} in a geometrically ambiguous neighborhood is inherently harder to classify than a point from an equally rare but geometrically distinctive class. This limitation motivates our geometry-aware extensions.

\subsection{Density-CB: Density Class Balanced Constraint}
\label{sec:ld_cb_loss}
Point cloud acquisition creates systematic density variations from sensor distance, angle, and occlusion, which can bias learning. To normalize for scan-dependent density variations, we introduce the Density-CB (Density Class Balanced) constraint. Unlike heuristic sampling, this constraint provides regularization based on local geometric density.

We first pre-compute the average local point density for each class. For every point \(i\), its local density is the number of neighbors within a fixed radius \(r\):
\begin{equation}
\label{eq:local_density}
d_i = \left|\{j : \|p_i - p_j\| \leq r\}\right|
\end{equation}
where \(\|\cdot\|\) denotes the Euclidean norm. This density quantifies the geometric context richness at point \(i\).

Radius selection. We use $r=0.2$~m because it matches the local scale of small MEP fittings while remaining sufficiently local relative to meter-scale pipes and beams. At this radius, the density estimate captures neighborhood occupancy around valves, elbows, tees, and flanges without merging large portions of adjacent structures into the same statistic. This choice is supported by the ablation in \autoref{tab:ablation_radius}, where $r=0.2$~m yields the strongest overall and tail-class performance; larger radii dilute the density signal by incorporating irrelevant points.

We then compute the class-level average local density:
\begin{equation}
\label{eq:class_avg_density}
\bar{d}_c = \frac{1}{n_c} \sum_{i: y_i = c} d_i
\end{equation}
The Density-CB loss modulates the standard CB weight \(w_c\) with a density-aware factor:
\begin{equation}
\label{eq:ld_cb_loss}
\mathcal{L}_{\text{Density-CB}} = \frac{1}{N} \sum_{i=1}^{N} w_{y_i} \cdot \gamma(\bar{d}_{y_i}) \cdot \phi_i \cdot \ell_{\text{CE}}(f_\theta(p_i), y_i)
\end{equation}
where the density modulator is:
\begin{equation}
\label{eq:density_modulator}
\gamma(d) = \frac{1}{1 + \log(d)}
\end{equation}
This function down-weights dense classes and up-weights sparse ones. The logarithmic scale ensures stability. The factor \(\phi_i = (1 - f_\theta(p_i)_{y_i})^{\gamma_{\mathrm{f}}}\), where \(f_\theta(p_i)_{y_i}\) is the predicted probability of the true class, is the focal factor of the CB+Focal baseline with \(\gamma_{\mathrm{f}}=2\); keeping it means that Density-CB differs from CB+Focal only in the density term. Density-CB requires a one-time \(\mathcal{O}(N \log N)\) pre-processing step and adds no training overhead.

\subsection{Boundary-CB: Boundary Class Balanced Constraint}
\label{sec:bc_cb_loss}
To resolve the geometric ambiguity central to the dual crisis, we propose the Boundary-CB (Boundary Class Balanced) constraint. This spatial context constraint targets point-level difficulty from confusing local contexts: part-to-part boundaries (compositional ambiguity) and regions where tail classes share primitives with head classes (primitive-similarity ambiguity). Both mechanisms create regions where network predictions exhibit high uncertainty.

The core innovation of Boundary-CB is using spatial context—specifically, neighborhood prediction consistency—to identify and emphasize geometrically ambiguous regions. Unlike Focal Loss, which relies on individual point confidence, Boundary-CB exploits 3D spatial structure by evaluating prediction consistency across local neighborhoods. This neighborhood-based approach addresses both compositional and primitive-similarity ambiguity simultaneously, as both manifest as regions of high prediction disagreement among spatially proximate points.

We quantify local consistency using Shannon entropy. For each point \(i\), let \(\mathcal{N}_k(i)\) be the indices of its \(k\)-nearest neighbors. We compute the average predicted probability distribution across this neighborhood:
\begin{equation}
\label{eq:neighborhood_prob}
\hat{p}_c^{(i)} = \frac{1}{k} \sum_{j \in \mathcal{N}_k(i)} f_\theta(p_j)_c, \quad c \in \{1, \ldots, C\}
\end{equation}
where \(f_\theta(p_j)_c\) is the predicted probability for point \(j\) being in class \(c\). The neighborhood entropy, normalized by its maximum \(\log C\), is:
\begin{equation}
\label{eq:neighborhood_entropy}
H_i = -\frac{1}{\log C} \sum_{c=1}^{C} \hat{p}_c^{(i)} \log \hat{p}_c^{(i)}
\end{equation}
Intuition. Consider a pipe-reducer transition. Points near the taper still look locally cylindrical, so neighboring predictions disagree: some neighbors vote for \texttt{Pipe}, while others vote for \texttt{Reducer}. This disagreement produces high entropy. In contrast, points well inside a uniform pipe region produce near-consistent neighborhood predictions and therefore low entropy.

From an information-theoretic perspective, the neighborhood entropy in Eq.~\ref{eq:neighborhood_entropy} measures local uncertainty caused by overlapping primitive geometry. In primitive-sharing regions, even a well-trained classifier can produce mixed neighborhood probabilities because adjacent points from different functional components have genuinely similar local features; high entropy is therefore treated as a structural signal of geometric ambiguity rather than merely a training artifact. This interpretation is consistent by analogy with classical Markov-random-field/Gibbs spatial regularization, where neighboring labels are coupled and boundary locations are precisely where homogeneous-region assumptions are violated~\cite{geman1984stochastic}. We use Shannon entropy rather than Gini impurity because it is smooth over the probability simplex, treats all uncertain class directions symmetrically, and does not require a predefined confusion-pair prior.

High entropy \(H_i\) indicates a geometrically ambiguous region where neighboring points have inconsistent predictions, which signals a challenging boundary between classes. The Boundary-CB constraint adaptively up-weights these points:
\begin{equation}
\label{eq:bc_cb_loss}
\mathcal{L}_{\text{Boundary-CB}} = \frac{1}{N} \sum_{i=1}^{N} w_{y_i} \cdot (1 + \alpha H_i) \cdot \phi_i \cdot \ell_{\text{CE}}(f_\theta(p_i), y_i)
\end{equation}
where \(\alpha > 0\) controls modulation strength and \(\phi_i\) is the focal factor defined in \S\ref{sec:ld_cb_loss}, so Boundary-CB also differs from CB+Focal only in its spatial term. This formulation directs model attention toward geometrically ambiguous regions during training.

Modulation strength \(\alpha\). Because Eq.~\ref{eq:neighborhood_entropy} divides by \(\log C\), the neighborhood entropy \(H_i\) lies in \([0, 1]\). With \(\alpha=1.0\), the modulator \((1 + \alpha H_i)\) therefore ranges from 1 (certain neighborhood) to 2 (maximally ambiguous boundary), providing emphasis on ambiguous regions without destabilizing training. We fix \(\alpha=1.0\) in all experiments and did not tune it.

Note on combined constraints. A multiplicative combination of Density-CB and Boundary-CB did not yield further performance improvements in our experiments. We hypothesize that compounding three re-weighting factors may over-regularize learning. We therefore recommend selecting either Density-CB (for density-dominated scenarios) or Boundary-CB (for geometric ambiguity-dominated scenarios such as the evaluated MEP scenes) based on dataset characteristics. See Supplementary Material for details.

\begin{figure*}[htbp]
    \centering
    \includegraphics[width=0.95\linewidth]{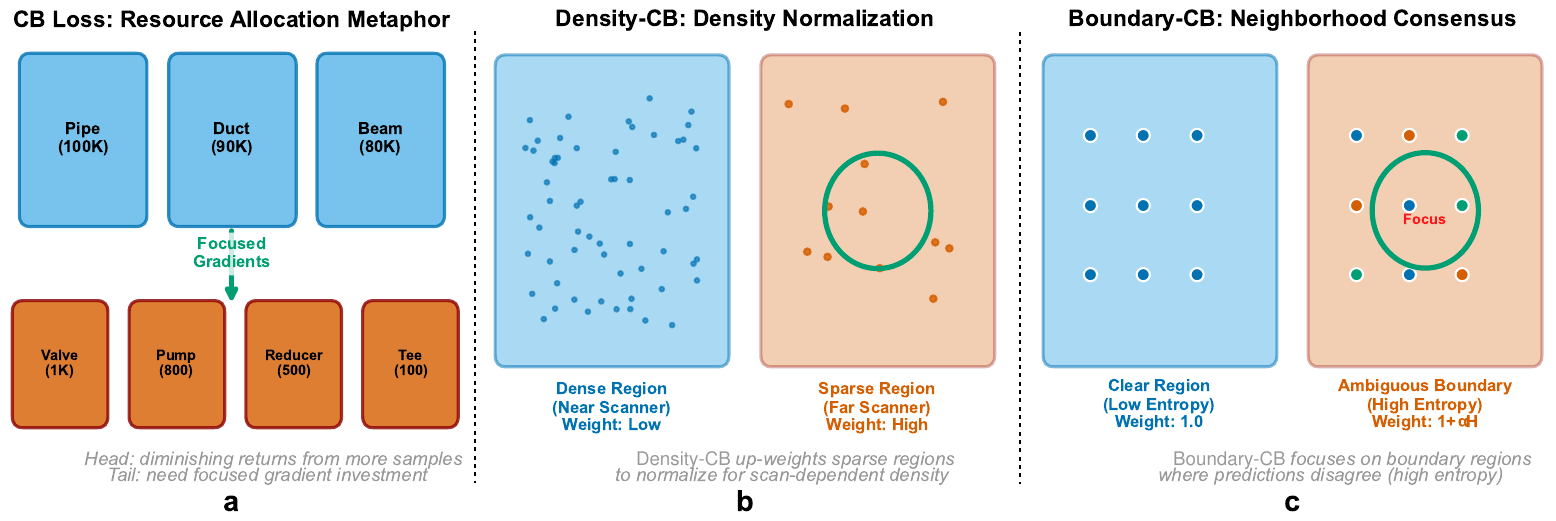}
    \caption{Intuitive framework for understanding spatial context constraints. This figure provides accessible conceptual understanding of how our three proposed methods resolve primitive-sharing ambiguity through intuitive metaphors. Panel A (CB Loss) shows the resource allocation metaphor: head classes (large departments: Pipe, Duct, Beam) receive diminishing returns from additional samples due to information overlap, while tail classes (small components: Valve, Pump, Reducer, Tee) benefit from focused gradient investment. The green arrow indicates that CB Loss directs learning resources where they are most needed, correcting statistical imbalance while preventing extreme re-weighting. Panel B (Density-CB) shows the density normalization metaphor: dense regions (near scanner, blue) contain many points and receive lower weights, while sparse regions (far from scanner, red) contain fewer points and are magnified for focused learning. The green circle highlights the magnifying effect on sparse tail classes that are spatially under-sampled. Panel C (Boundary-CB) shows the neighborhood consensus metaphor: clear regions (uniform blue points, low entropy) receive baseline weights where predictions are consistent, while ambiguous boundaries (mixed colors, high entropy) are up-weighted to resolve uncertainty. The green circle indicates focused learning on challenging transition zones where primitive-sharing creates geometric confusion (e.g., pipe-reducer boundaries). Together, these metaphors illustrate how each constraint addresses a distinct challenge: CB corrects statistical imbalance, Density-CB normalizes spatial sampling bias, and Boundary-CB resolves geometric ambiguity through spatial context.}
    \label{fig:loss_intuition_metaphor}
\end{figure*}

\begin{figure*}[htbp]
    \centering
    \includegraphics[width=0.95\linewidth]{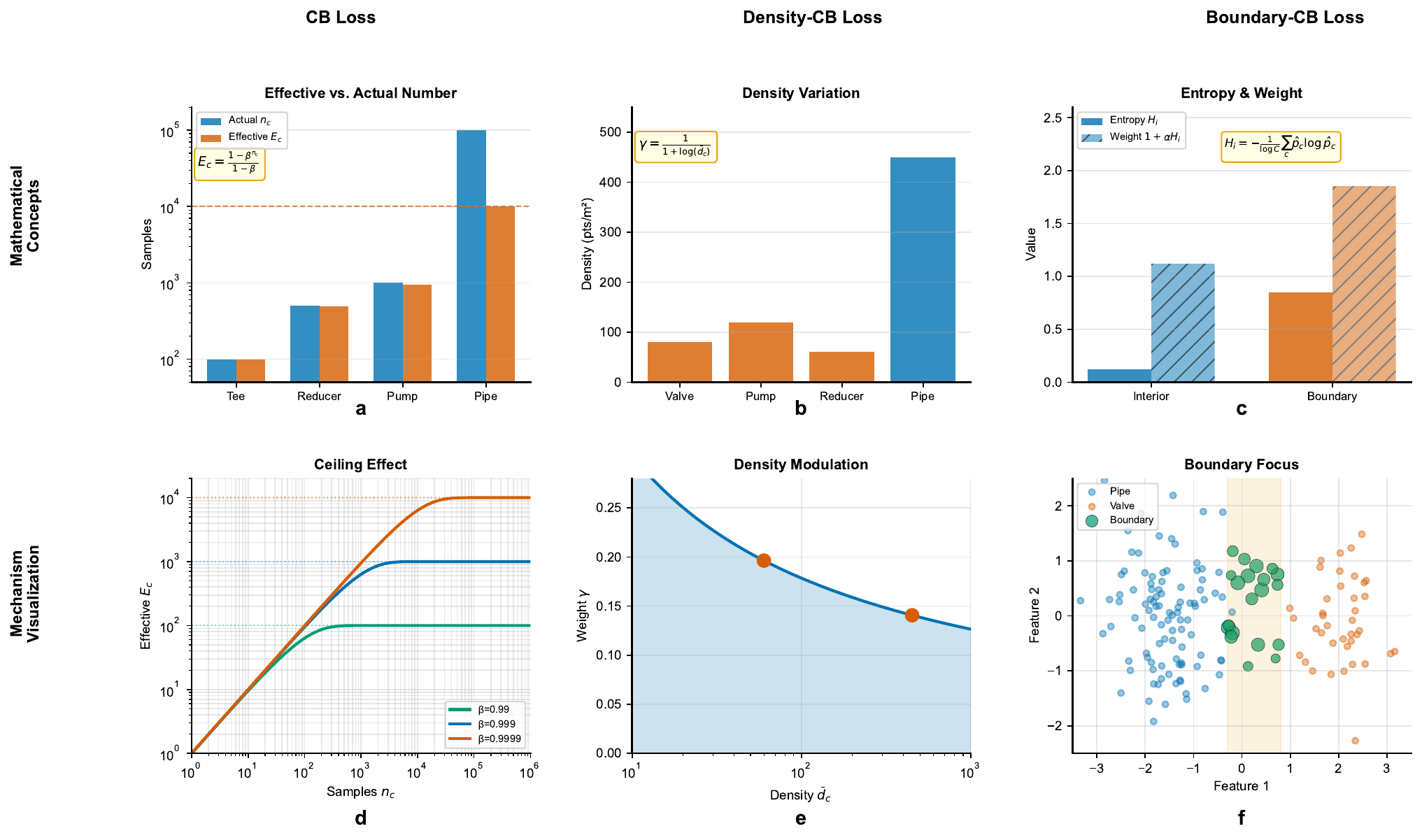}
    \caption{Mathematical foundations and mechanisms for resolving primitive-sharing ambiguity. This figure formalizes how our spatial context constraints address geometric ambiguity through three complementary mechanisms. The top row shows mathematical concepts: Panel A shows the effective number concept $E_c = (1-\beta^{n_c})/(1-\beta)$, where CB Loss corrects statistical imbalance by quantifying diminishing marginal information from additional samples; Panel B visualizes scan-dependent density variation $\gamma(\bar{d}_c) = 1/(1+\log \bar{d}_c)$, where Density-CB normalizes spatial sampling bias by down-weighting dense regions and up-weighting sparse ones; Panel C illustrates the normalized neighborhood entropy $H_i = -\frac{1}{\log C}\sum_c \hat{p}_c^{(i)} \log \hat{p}_c^{(i)}$ and the resulting weight $1+\alpha H_i$, where Boundary-CB quantifies prediction inconsistency to identify geometrically ambiguous boundaries. The bottom row shows mechanism visualizations: Panel D demonstrates the ceiling effect across $\beta$ values, showing how CB prevents extreme re-weighting while still emphasizing rare classes; Panel E plots the density modulator curve, revealing logarithmic down-weighting of dense head classes (pipes, beams) and up-weighting of sparse tail classes (valves, reducers); Panel F visualizes how Boundary-CB identifies ambiguous boundaries in feature space, focusing learning on transition zones (green) where spatial context is most critical for resolving primitive-sharing confusion. Together, these mechanisms demonstrate how spatial context constraints extend frequency-based re-weighting to resolve geometric ambiguity.}
    \label{fig:loss_mechanism}
\end{figure*}

To enhance methodological transparency, we provide two complementary interpretability visualizations. \autoref{fig:loss_intuition_metaphor} offers accessible conceptual intuition through vivid real-world metaphors (resource allocation for CB Loss, magnifying glass for Density-CB, neighborhood consensus for Boundary-CB), bridging the gap between mathematical formalism and intuitive understanding. \autoref{fig:loss_mechanism} presents the corresponding mathematical foundations and operational mechanisms, formalizing how each constraint addresses statistical imbalance, density bias, and geometric ambiguity through precise equations and visual demonstrations. Together, these figures provide both an intuitive entry point and a rigorous explanation of how neighborhood-based prediction consistency resolves primitive-sharing ambiguity.

\subsection{Evaluation Metrics}
\label{sec:evaluation_metrics}
We use standard and specialized metrics to evaluate long-tailed performance.

Traditional metrics. For \(C\) classes, let \(TP_c\), \(FP_c\), \(FN_c\), and \(TN_c\) denote true positives, false positives, false negatives, and true negatives for class \(c\). The Intersection-over-Union (IoU) for class \(c\) is:
\begin{equation}
\label{eq:iou}
\text{IoU}_c = \frac{TP_c}{TP_c + FP_c + FN_c}
\end{equation}
\begin{equation}
\label{eq:miou}
\text{mIoU} = \frac{1}{C} \sum_{c=1}^{C} \text{IoU}_c
\end{equation}
Overall accuracy (OA) measures the proportion of correctly classified points. While useful, these metrics can be dominated by high accuracy on abundant head classes.

Long-tailed specific metrics. To assess performance on the dual crisis, we partition classes by frequency. Following the distribution of Industrial3D, we define three groups: \(\mathcal{C}_{\text{head}}\) (3 most frequent classes, 77\% of points), \(\mathcal{C}_{\text{common}}\) (2 moderately frequent classes), and \(\mathcal{C}_{\text{tail}}\) (7 rarest classes, each <3\% of points). We then compute mIoU for each group (\(\text{mIoU}_{\text{head}}\), \(\text{mIoU}_{\text{common}}\), \(\text{mIoU}_{\text{tail}}\)). The tail-class mIoU directly measures performance on critical rare classes. To evaluate balance, we compute the Harmonic Mean IoU (H-IoU):
\begin{equation}
\label{eq:hiou}
\text{H-IoU} = \frac{2 \cdot \text{mIoU}_{\text{head}} \cdot \text{mIoU}_{\text{tail}}}{\text{mIoU}_{\text{head}} + \text{mIoU}_{\text{tail}}}
\end{equation}
The harmonic mean penalizes large disparities between head and tail performance. A high H-IoU indicates that tail-class improvements have been achieved without degrading head-class accuracy.

\section{Experiments}
\label{sec:experiments}
This section evaluates whether spatial context constraints reduce primitive-sharing errors on the predefined Industrial3D split. We present experimental setup (\S\ref{sec:exp_setup}), main quantitative results (\S\ref{sec:main_results}), ablation studies (\S\ref{sec:ablation_studies}), and qualitative analysis (\S\ref{sec:qualitative_analysis}).

\subsection{Experimental Setup}
\label{sec:exp_setup}

\subsubsection{Datasets and Class Distribution}
\begin{figure}[!htbp]
    \centering
    \includegraphics[width=0.95\linewidth]{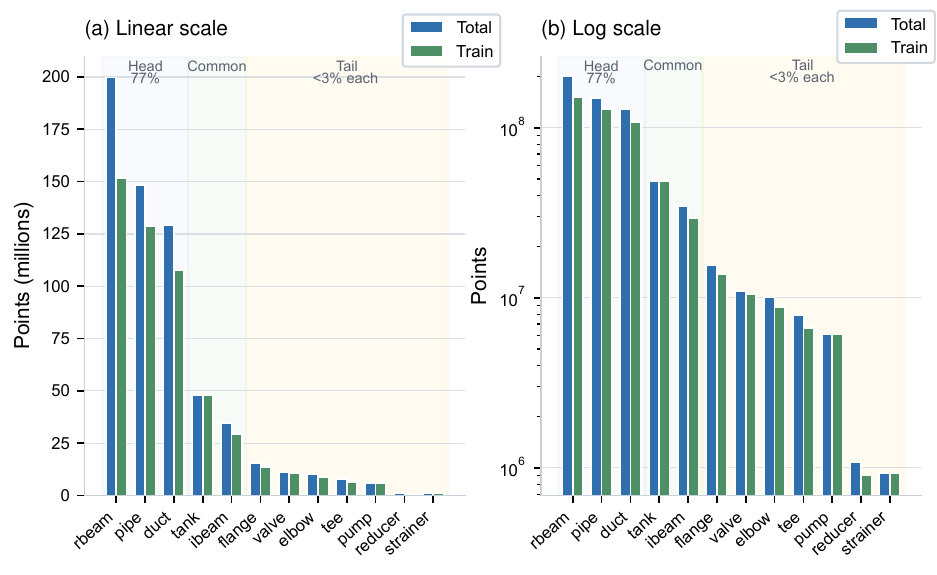}
    \caption{Class distribution in the Industrial3D dataset. The 12 classes are grouped by frequency: Head classes (Duct, Pipe, RectangularBeam) comprise 77\% of the 612.7M labelled points; Common classes (Ibeam, Tank) have moderate representation; Tail classes (Flange, Elbow, etc.) each represent <3\% of points. This grouping reflects the dual crisis: head classes are uniformly long-shaped structures, while tail classes are geometrically diverse, including complex assemblies (composite-tail) and simple primitives (primitive-similarity tail). The severe imbalance ($\sim$215:1 ratio) necessitates advanced long-tailed learning approaches.}
    \label{fig:distribution}
\end{figure}

We evaluate on the Industrial3D dataset~\cite{yin2026industrial3d}, a large-scale TLS-based MEP point cloud dataset from water-treatment facilities (20 room scenes grouped into 13 areas across 7 facilities, 612.7M labelled points, 12 classes). Compared with object-centric industrial 3D benchmarks such as MVTec ITODD~\cite{drost2017itodd}, which targets pose estimation of isolated parts on simple backgrounds, Industrial3D provides registered facility-scale MEP scenes with primitive-sharing tail classes that drive the dual crisis studied here. Industrial3D exemplifies the dual crisis with its combination of extreme class imbalance and high geometric ambiguity. The key properties of Industrial3D are:
\begin{itemize}
    \item Statistical bias: A severely long-tailed distribution (>200:1 imbalance ratio). Classes are grouped into Head (3 classes, 77\% of points), Common (2 classes), and Tail (7 classes, each <3\% of points), as shown in \autoref{fig:distribution}.
    \item Geometric ambiguity: Head classes are long-shaped continuous structures (beams, pipes, ducts), while tail classes exhibit two distinct patterns: 
    \begin{itemize}
        \item \emph{Composite-tail} (\texttt{Flange}, \texttt{Valve}, \texttt{Pump}, \texttt{Strainer}): Multi-part assemblies more structurally complex than head classes, yet composed of head-class-like primitives, leading to boundary confusion.
        \item \emph{Primitive-similarity tail} (\texttt{Elbow}, \texttt{Tee}, \texttt{Reducer}): Simple objects sharing nearly identical cylindrical primitives with head-class \texttt{Pipe}, making them locally indistinguishable without global context.
    \end{itemize}
    This structural pattern means all tail classes suffer from geometric confusion with dominant head-class shapes.
\end{itemize}

Industrial3D was captured in operational water-treatment facilities using a Leica BLK360 terrestrial laser scanner with approximately 6~mm point spacing at 10~m. Five domain experts annotated the registered point clouds in CloudCompare through a tiling, cluster extraction, merging, and boundary-refinement workflow, requiring approximately 754 person-hours. We use the predefined area-based benchmark split to avoid direct spatial overlap: 11 areas are used for training (527.8M points, 86.1\%), while Area~6 and Area~12 form the evaluation split (84.9M points, 13.9\%). Area~6 comprises Pumping Station Unit and Variable Pump Area rooms dominated by rectangular beams (54.5\% of its points) and ducts (27.0\%); Area~12 is a Sludge Press House containing all 12 labelled classes and substantially more mechanical equipment. The two areas therefore differ strongly in function and class composition: approximately 1.55\% of labelled Area~6 points belong to tail classes, compared with approximately 20.15\% in Area~12. The companion benchmark selected these two areas as spatially distinct water-treatment subsystems with complementary characteristics~\cite{yin2026industrial3d}, and we retained the split for comparability. We did not run $k$-fold validation. Unlike S3DIS, where most classes occur in every area, Industrial3D's tail classes are concentrated in a few areas: Strainer occurs in only four of the 13 areas (Areas~1, 7, 9, and~12), Area~1 alone holds 63.8\% of all Strainer points, and Area~12 is the only area that contains all 12 classes. Folds would therefore have to be grouped by hand so that each test fold contains every class, and a fold that holds out Area~1 would train on only about one third of the Strainer points, so fold-to-fold differences in tail-class IoU would partly reflect the training data left in each fold. Such grouped folds exist (for example, Areas \{1, 4, 11\}, \{2, 7, 8, 13\}, \{3, 5, 9, 10\}, and \{6, 12\} each contain all 12 classes), but evaluating them requires retraining every compared model on each fold, which we leave to a future study with all settings fixed in advance. Because the predefined split has no separate validation area, the evaluation areas also served as the validation split: for every method, including the baseline, we kept the checkpoint with the best evaluation-split mIoU, and the $k$ and $r$ sweeps were run on this split. Every tested $k$ nevertheless gives higher mIoU, tail mIoU, and H-IoU than the baseline (\autoref{tab:main_results} and \autoref{tab:ablation_bc_k}), so the gain does not depend on choosing the best $k$. We therefore interpret the reported scores as within-dataset development-split evidence rather than an independent estimate of cross-dataset generalization. Because a pooled score can hide differences between the two areas, \autoref{tab:per_area_results} reports Areas~6 and~12 separately. The complete area--class inventory appears in the Supplementary Material.

\begin{table*}[!htbp]
\centering
\caption{Area-level composition of Industrial3D. Counts are derived from the labelled-point inventory and rounded to the nearest 0.1M for display. ``Classes'' is the number of the 12 benchmark classes present in the area; ``Tail'' is the within-area percentage of Flange, Valve, Pump, Strainer, Elbow, Tee, and Reducer points. Areas 6 and 12 form the predefined evaluation split. Full class counts and percentages for every area are reported in the area-level composition tables in the Supplementary Material.}
\label{tab:area_composition_main}
\resizebox{\textwidth}{!}{%
\begin{tabular}{cllrrr|cllrrr}
\toprule
Area & Function & Split & Points (M) & Classes & Tail (\%) & Area & Function & Split & Points (M) & Classes & Tail (\%) \\
\midrule
1 & Chiller & Train & 47.0 & 11 & 26.97 & 8 & VPA & Train & 69.4 & 11 & 3.89 \\
2 & Service Gallery & Train & 79.6 & 10 & 7.80 & 9 & OSCG & Train & 15.1 & 9 & 6.60 \\
3 & Tank & Train & 61.9 & 8 & 2.35 & 10 & PBF-1 & Train & 52.1 & 8 & 18.02 \\
4 & Corridor & Train & 52.8 & 8 & 0.50 & 11 & PBF-2 & Train & 50.2 & 8 & 5.74 \\
5 & Ozone Generation & Train & 48.8 & 8 & 4.26 & \textbf{12} & \textbf{Sludge Press House} & \textbf{Evaluation} & \textbf{37.5} & \textbf{12} & \textbf{20.15} \\
\textbf{6} & \textbf{PSU/VPA} & \textbf{Evaluation} & \textbf{47.4} & \textbf{8} & \textbf{1.55} & 13 & WRT & Train & 20.0 & 10 & 22.33 \\
7 & Oxygen Compressor/Instrument & Train & 30.8 & 9 & 3.65 & & & & & & \\
\bottomrule
\end{tabular}%
}
\end{table*}

\subsubsection{Compared Methods}
All methods use ResPointNet++~\cite{yin2021} as the backbone. We compare:
\begin{itemize}
    \item Baseline: Standard cross-entropy loss.
    \item CB Loss: Class-Balanced Loss~\cite{cui2019class} with frequency-based re-weighting.
    \item CB+Focal Loss: CB Loss with focal loss modulation, a strong frequency-based baseline.
    \item Density-CB (Ours): Density Class Balanced constraint that adds density normalization to CB+Focal.
    \item Boundary-CB (Ours): Boundary Class Balanced constraint that adds spatial context to CB+Focal for resolving geometric ambiguity.
    \item Combined (Ours): Both Density-CB and Boundary-CB constraints applied jointly.
\end{itemize}

The within-backbone comparison is designed to isolate the effect of the proposed loss-level constraints: all methods use the same ResPointNet++ backbone, data split, optimizer, and augmentation pipeline. Baselines were selected to represent the major method families relevant to the long-tailed learning taxonomy discussed in Section~\ref{sec:long_tailed}. Standard cross-entropy establishes the backbone baseline; CB and CB+Focal represent frequency-based loss modification; and Density-CB, Boundary-CB, and the combined variant test whether spatial density or neighborhood uncertainty adds value beyond class frequency alone. The selection criteria were applicability to dense per-point semantic segmentation, reproducible implementation or benchmark availability, and representativeness of distinct training-objective families.

To broaden empirical context in this revision, we conducted two additional same-dataset experiments with modern architectures, RandLA-Net and Point Transformer v3 (PTv3), under the unified Industrial3D setting~\cite{yin2026industrial3d}. As summarized in \autoref{tab:industrial3d_backbone_context}, RandLA-Net reaches 39.83\% mIoU, PTv3 reaches 41.90\% mIoU, and ResPointNet++ reaches the highest mIoU among the three backbones at 52.48\%. We therefore select ResPointNet++ as the backbone for the controlled Boundary-CB and Density-CB experiments, so the proposed loss-level constraints can be evaluated on a competitive domain-established backbone while keeping the architecture fixed.

\begin{table}[!htbp]
    \centering
    \caption{
    Fully supervised same-dataset Industrial3D backbone results on the predefined Industrial3D evaluation split (Areas~6 and~12; mIoU and per-class IoU, \%). Bold indicates the best value in each metric column; ties are bolded. Abbreviations: Dct=Duct, Elb=Elbow, Flg=Flange, Ibm=I-beam, Pmp=Pump, Rbm=RectangularBeam, Rdr=Reducer, Str=Strainer, and Val=Valve.}

    \label{tab:industrial3d_backbone_context}
    \resizebox{\textwidth}{!}{
    \begin{tabular}{lccccccccccccc}
    \hline
    
    \textbf{Method} & \textbf{mIoU} & \textbf{Dct} & \textbf{Elb} & \textbf{Flg} & \textbf{Ibm} & \textbf{Pipe} & \textbf{Pmp} & \textbf{Rbm} & \textbf{Rdr} & \textbf{Str} & \textbf{Tank} & \textbf{Tee} & \textbf{Val} \\
    \hline
    RandLA-Net~\citep{hu2020randlanet} & 39.83 & 83.37 & 18.06 & 23.92 & 81.14 & 68.97 & 0.98 & 91.25 & 0.00 & 0.00 & 71.89 & 3.03 & 35.36 \\
    PTv3~\citep{wu2024point} & 41.90 & \textbf{97.39} & \textbf{55.40} & 33.35 & \textbf{99.71} & \textbf{91.80} & 0.00 & \textbf{98.53} & 0.00 & 0.00 & 0.00 & \textbf{26.61} & 0.00 \\
    ResPointNet++~\citep{yin2021} & \textbf{52.48} & 90.73 & 40.77 & \textbf{45.23} & 98.66 & 76.33 & \textbf{43.08} & 94.79 & 0.00 & 0.00 & \textbf{98.99} & 3.10 & \textbf{38.06} \\
    \hline
    \end{tabular}
    }
    \vspace{0.1in}
\end{table}

Note on classic baselines. We do not include inverse frequency (IF) or square-root inverse frequency (SRIF) weighting in our main comparison. For Industrial3D's extreme 215:1 imbalance, IF leads to training instability. As shown by Cui et al.~\cite{cui2019class}, CB Loss provides superior stability and performance over naive IF/SRIF on severely imbalanced datasets. We therefore focus on CB-based methods, which represent the state-of-the-art for this problem.

\subsubsection{Implementation Details}
Models were implemented in PyTorch and trained on two NVIDIA RTX 3090 GPUs with stochastic gradient descent (momentum 0.98, weight decay $10^{-3}$) and an initial learning rate of 0.01, decayed by a factor of 0.9885 per epoch. The baseline was trained for 215 epochs and CB+Focal, Density-CB, and Boundary-CB for 262 epochs; for every run we used the checkpoint with the best evaluation-split mIoU (epoch 165 for the baseline and 212 for the other three). Data augmentation included random rotation, scaling, and jittering. Hyperparameters were set as follows: $\beta=0.9999$ for the CB component; for Density-CB, radius $r=0.2$ m; for Boundary-CB, we evaluated neighborhood sizes $k\in\{8,16,32,64,128\}$ with modulation strength $\alpha=1.0$; $k=64$ gave the best tested result.

For large-scene training, we follow the ResPointNet++ large-scale segmentation protocol~\cite{yin2021}. Registered scenes are queried as spherical sub-clouds with a 2~m radius; training spheres are selected randomly, whereas test spheres are selected regularly to cover the scene. Points are first grid-subsampled with a 4~cm base grid, where one point is randomly retained per occupied grid cell, and a fixed 16,384 points are sampled from each sub-cloud as network input. The hierarchical encoder follows the same grid-subsampled representation, with progressively coarser resolutions and ball-query neighborhoods. Thus, the 612.7M labelled points are processed as many local sub-clouds rather than as one full scene in GPU memory. Density-CB uses this same grid-subsampled coordinate representation for local density estimation, so the density term reflects the effective spatial density seen by the network after preprocessing rather than an additional class-aware over- or under-sampling scheme.

\subsection{Main Results}
\label{sec:main_results}
\autoref{tab:main_results} presents the primary comparison. On the predefined evaluation split, spatial context constraints reduce errors associated with geometric ambiguity, with Boundary-CB achieving better performance than frequency-only baselines.

\begin{figure}[!htbp]
\centering
\includegraphics[width=\linewidth]{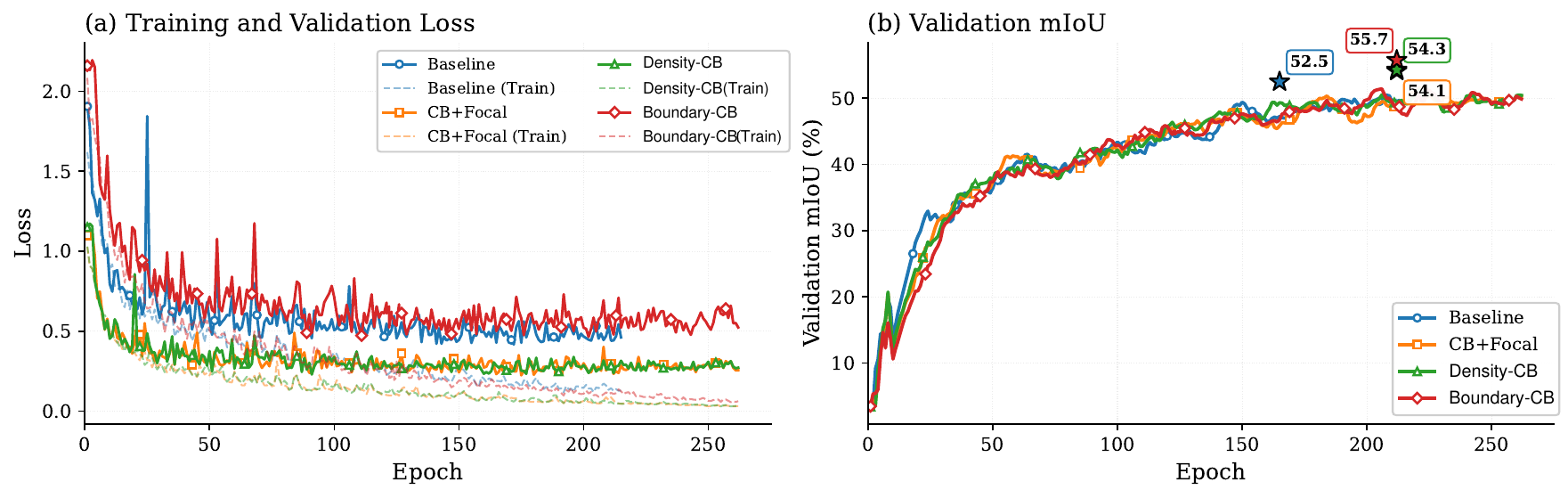}
\caption{Training dynamics comparison across different methods on the Industrial3D dataset.
(a) Training and development-split evaluation loss curves (215 epochs for the baseline and 262 for the other models). Solid lines represent evaluation loss, while dashed lines show training loss. All methods converge; because each loss weights points differently, the loss values are not directly comparable across methods.
(b) Development-split evaluation mIoU curves computed during training with two test-time votes (smoothed with window size 10 for clarity). Star markers indicate the single-epoch peaks reported in \autoref{tab:main_results}: Boundary-CB 55.7\%, Baseline 52.5\%, CB+Focal 54.1\%, and Density-CB 54.3\%. The smoothed curves of the four methods overlap in later epochs, so the differences in \autoref{tab:main_results} rest on the selected checkpoints; a complete 20-vote evaluation of these checkpoints (\autoref{tab:per_area_results}) gives 55.62\% for Boundary-CB and 52.66\% for the baseline. These curves are not an independent generalization estimate. In the figure, ''Validation'' denotes the Area~6+12 evaluation split.}
\label{fig:curve_loss_miou}
\end{figure}

Overall performance. Boundary-CB ($k=64$) achieves 55.74\% mIoU, a +1.65 pp gain over CB+Focal. Tail-class mIoU reaches 29.59\% (+5.27 pp; +21.7\% relative vs.\ baseline), showing that spatial context adds useful information beyond class frequency in the evaluated MEP split. The harmonic mean IoU (H-IoU) improves from 38.04\% to 44.31\% (+16.5\% relative), demonstrating better head--tail balance. Head-class mIoU slightly improves (87.28\% to 88.14\%), so tail and H-IoU gains are achieved without sacrificing head performance on this split. The best tested result at $k=64$ shows that context scale affects disambiguation; the larger $k=128$ setting degrades performance by introducing noise. \autoref{fig:curve_loss_miou} shows the training dynamics; because each loss weights points differently, the loss values are not directly comparable across methods.

Per-class analysis. \autoref{tab:per_class_iou} shows per-class performance, with different gains across tail-class subgroups. Boundary-CB ($k=64$) dramatically improves previously unrecognizable classes: \texttt{Reducer} improves from 0\% (baseline) to 21.12\% IoU, while \texttt{Valve} reaches 47.32\% (+24.3\% relative vs. 38.06\% baseline). The verified neighborhood sweep also shows that classwise sensitivity is heterogeneous rather than monotonic: \texttt{Elbow} peaks at $k=32$, \texttt{Tee} attains its best Boundary-CB score at $k=8$, and \texttt{Reducer}, \texttt{Valve}, and \texttt{Pump} favor $k=64$. A classwise sensitivity table is provided in the Supplementary Material.

\begin{itemize}
    \item Composite-tail classes: Boundary-CB ($k=64$) shows gains on \texttt{Valve} (47.32\%) and \texttt{Pump} (44.37\%). Within the verified Boundary-CB sweep, \texttt{Flange} peaks earlier at $k=16$, showing that not all multi-part assemblies require the same context width.

    \item Primitive-similarity tail classes: Boundary-CB ($k=64$) achieves breakthrough performance on \texttt{Reducer} (21.12\% vs. 0\% baseline), showing that larger spatial context can capture subtle geometric signatures that separate reducers from dominant pipes. \texttt{Elbow} performs best at $k=32$, while \texttt{Tee} is unstable and achieves its best Boundary-CB score at $k=8$, suggesting that compact or branching geometries are more sensitive to over-contextualization.
\end{itemize}

The 0\% IoU on \texttt{Strainer} under all tested losses indicates an extreme few-shot scenario beyond the capabilities of loss re-weighting, requiring complementary techniques such as meta-learning or synthetic data augmentation.

Per-area analysis. \autoref{tab:per_area_results} evaluates the same four models on Areas~6 and~12 separately. Each area mean averages only the classes present in that area's ground truth, so Area~6 uses eight classes and Area~12 uses all 12. The table also reports the two rooms of Area~6 (Area~12 is a single room), and the Supplementary Material gives the classwise IoUs for each area. The gain is concentrated in Area~12, where Boundary-CB raises tail mIoU from 21.42\% (baseline) to 26.53\% and H-IoU from 33.43\% to 39.51\%, mainly through Reducer (0\% to 16.90\% IoU) and Valve (38.73\% to 46.11\%). Area~6 has 1.55\% tail points, no Pump, Reducer, Strainer, or Tank points, and only 559 Valve points, on which every model scores 0\%. There the four models are within 0.7 percentage points in mIoU, and Boundary-CB scores lowest (61.01\% vs.\ 61.65\% for the baseline; 0.2 and 1.4 points below it in rooms 6-1 and 6-2), mainly on Flange and Tee. Three factors bear on this result, and two evaluation areas cannot separate them. First, Area~6 is closer to the training data than Area~12: both of its rooms come from the facility that contributes training Areas~3, 4, 5, 7, and~8, and room 6-2 is one of six Variable Pump Area rooms whose other five are in training Area~8, whereas Area~12 is the only area from its facility. Second, the confusion matrices show that the Area~6 decrease is concentrated on two classes with little support there: Flange and Tee (131K and 163K points) account for about 70\% of it, mostly in room 6-2, where Boundary-CB labels about 14K more Pipe and Elbow points as Flange and about 3K fewer Tee points correctly. Class composition alone does not explain the difference, because Flange, Elbow, and Pipe gain 3.5--4.6 points in Area~12 but lose 0.4--2.2 points in Area~6. Third, checkpoints were selected on the pooled evaluation split, whose tail points lie almost entirely in Area~12 (all Pump, Reducer, Strainer, and Tank points and 99.97\% of Valve points), so the Area~12 gain may be optimistic. In Area~12, Boundary-CB labels about a quarter of the Reducer points correctly, whereas the baseline labels no point as Reducer. Boundary-CB also predicts Valve, Elbow, and Pump more precisely: the points wrongly labelled Valve fall from 1.48M to 0.68M, and Elbow precision rises from 42.5\% to 68.2\%. Pooled over both areas, this evaluation pass gives 52.66\% (baseline), 52.05\% (CB+Focal), 53.49\% (Density-CB), and 55.62\% (Boundary-CB) mIoU: Boundary-CB remains the best, whereas the gain of CB+Focal over the baseline in \autoref{tab:main_results} does not hold.

\begin{table*}[!htbp]
\centering
\caption{Main performance comparison on the predefined Industrial3D evaluation split. We report Overall mIoU, and group-wise mIoU for Head (3 classes), Common (2 classes), and Tail (7 classes) categories, alongside the Harmonic Mean IoU (H-IoU) to measure head--tail balance. All values are in percentage. Values are the best-epoch scores on this split recorded during training with two test-time votes, and the checkpoints were selected on the same split; \autoref{tab:per_area_results} reports a complete 20-vote evaluation of these checkpoints for each area. The proposed Boundary-CB spatial context constraint ($k=64$) achieves the best tested trade-off, reducing primitive-sharing errors while maintaining head-class accuracy on this split. See \autoref{sec:ablation_studies} for detailed ablation analysis linking these results to spatial context parameters.}
\label{tab:main_results}
\begin{tabular}{lccccc}
\toprule
Method & Overall & Head & Common & Tail & H-IoU \\
\midrule
ResPointNet++ (Baseline) & 52.48 & 87.28 & 98.82 & 24.32 & 38.04 \\
CB+Focal Loss & 54.09 & 87.39 & 99.06 & 26.97 & 41.21 \\
\midrule
Density-CB (Ours) & 54.27 & 88.05 & 98.28 & 27.23 & 41.59 \\
Boundary-CB (Ours, k=64) & \textbf{55.74} & \textbf{88.14} & 98.64 & \textbf{29.59} & \textbf{44.31} \\
Combined (Ours) & 53.59 & 87.05 & 97.38 & 26.74 & 40.91 \\
\bottomrule
\end{tabular}
\end{table*}

\begin{table*}[!htbp]
\centering
\caption{Per-class IoU comparison on Industrial3D. Comparison of baseline and frequency-based methods against our spatial context constraints. Boundary-CB ($k=64$) performs better on geometrically ambiguous tail classes, particularly \texttt{Reducer} (0\%$\rightarrow$21.12\%) and \texttt{Valve} (+24.3\% relative), showing that spatial context resolves primitive-sharing ambiguity. For brevity, we report the best-performing Boundary-CB variant ($k=64$); the effect of neighborhood size $k$ on component-specific trade-offs is analyzed in \autoref{sec:ablation_studies} and \autoref{sec:discussion}.}
\label{tab:per_class_iou}
\adjustbox{width=\textwidth}{%
\begin{tabular}{lcccccc}
\toprule
Class & Group & Baseline & CB+Focal & Density-CB & Boundary-CB (k=64) & Combined \\
\midrule
\texttt{Duct} & Head & 90.73 & 91.40 & \textbf{91.42} & 91.08 & 88.58 \\
\texttt{Pipe} & Head & 76.33 & 75.69 & 77.60 & 78.32 & \textbf{78.76} \\
\texttt{RectangularBeam} & Head & 94.79 & 95.08 & \textbf{95.13} & 95.02 & 93.82 \\
\midrule
\texttt{Ibeam} & Common & 98.66 & \textbf{99.08} & 99.03 & 98.51 & 98.58 \\
\texttt{Tank} & Common & 98.99 & \textbf{99.04} & 97.53 & 98.76 & 96.19 \\
\midrule
\multicolumn{7}{l}{\textit{Composite-Tail (complex multi-part assemblies)}} \\
\texttt{Flange} & Tail & 45.23 & \textbf{50.71} & 48.96 & 47.85 & 47.57 \\
\texttt{Valve} & Tail & 38.06 & 32.76 & 39.45 & \textbf{47.32} & 45.16 \\
\texttt{Pump} & Tail & 43.08 & 43.81 & \textbf{47.01} & 44.37 & 35.58 \\
\texttt{Strainer} & Tail & 0.00 & 0.00 & 0.00 & 0.00 & 0.00 \\
\multicolumn{7}{l}{\textit{Primitive-Similarity Tail (cylindrical, pipe-like geometry)}} \\
\texttt{Elbow} & Tail & 40.77 & 41.88 & 40.58 & 43.23 & \textbf{44.52} \\
\texttt{Tee} & Tail & 3.10 & 4.68 & \textbf{5.75} & 3.24 & 4.52 \\
\texttt{Reducer} & Tail & 0.00 & 14.92 & 8.83 & \textbf{21.12} & 9.83 \\
\bottomrule
\end{tabular}%
}
\end{table*}

\begin{table*}[!htbp]
\centering
\caption{Results for each evaluation area and room (\%). All four models use the checkpoints reported in \autoref{tab:main_results}. mIoU and the Head, Common, and Tail means average only the classes present in the ground truth of the area or room. Area~6 contains eight of the 12 classes: its Common group is Ibeam only, and its Tail group is Elbow, Flange, Tee, and Valve (Valve has only 559 labelled points). It consists of room 6-1 (Pumping Station Unit), which has no Ibeam points and contains all the Valve points, and room 6-2 (Variable Pump Area), which has no Valve points. Area~12 (Sludge Press House) is a single room and contains all 12 classes. H-IoU combines the Head and Tail means as in \autoref{eq:hiou}. All values come from one complete evaluation pass with 20 test-time votes. Pooled over both areas, this pass gives 52.66\%, 52.05\%, 53.49\%, and 55.62\% mIoU (tail mIoU 24.36\%, 23.59\%, 25.73\%, and 29.23\%) for the baseline, CB+Focal, Density-CB, and Boundary-CB. \autoref{tab:main_results} instead reports the best-epoch evaluation-split score logged during training with two votes per epoch; this pass reproduces its mIoU within 0.2 percentage points for the baseline and Boundary-CB, but gives 0.8 points less for Density-CB and 2.0 points less for CB+Focal. Bold marks the best value in each column within an area or room.}
\label{tab:per_area_results}
\adjustbox{max width=\textwidth}{%
\begin{tabular}{llccccc}
\toprule
Area/room & Method & mIoU & Head & Common & Tail & H-IoU \\
\midrule
Area 6 (8 classes) & ResPointNet++ (Baseline) & \textbf{61.65} & 97.56 & 99.63 & \textbf{25.23} & \textbf{40.09} \\
 & CB+Focal Loss & 61.49 & \textbf{97.62} & \textbf{99.80} & 24.80 & 39.56 \\
 & Density-CB (Ours) & 61.53 & 97.42 & 99.69 & 25.07 & 39.88 \\
 & Boundary-CB (Ours, k=64) & 61.01 & 97.35 & 99.58 & 24.12 & 38.66 \\
\midrule
Area 12 (12 classes) & ResPointNet++ (Baseline) & 48.01 & 76.11 & 98.94 & 21.42 & 33.43 \\
 & CB+Focal Loss & 47.57 & 75.45 & \textbf{98.95} & 20.94 & 32.78 \\
 & Density-CB (Ours) & 49.02 & 76.94 & 98.61 & 22.88 & 35.28 \\
 & Boundary-CB (Ours, k=64) & \textbf{51.25} & \textbf{77.34} & 98.66 & \textbf{26.53} & \textbf{39.51} \\
\midrule
Room 6-1 (7 classes) & ResPointNet++ (Baseline) & \textbf{51.86} & 97.48 & N/A & \textbf{17.64} & \textbf{29.87} \\
 & CB+Focal Loss & 51.53 & \textbf{97.55} & N/A & 17.02 & 28.98 \\
 & Density-CB (Ours) & 50.16 & 97.14 & N/A & 14.92 & 25.86 \\
 & Boundary-CB (Ours, k=64) & 51.63 & 97.27 & N/A & 17.41 & 29.53 \\
\midrule
Room 6-2 (7 classes) & ResPointNet++ (Baseline) & 71.54 & 97.60 & 99.63 & 36.11 & 52.71 \\
 & CB+Focal Loss & 71.07 & \textbf{97.67} & \textbf{99.80} & 34.89 & 51.41 \\
 & Density-CB (Ours) & \textbf{71.55} & 97.59 & 99.69 & \textbf{36.14} & \textbf{52.74} \\
 & Boundary-CB (Ours, k=64) & 70.15 & 97.46 & 99.58 & 33.03 & 49.34 \\
\bottomrule
\end{tabular}%
}
\end{table*}

\begin{figure}[!htbp]
    \centering
    \includegraphics[width=0.95\linewidth]{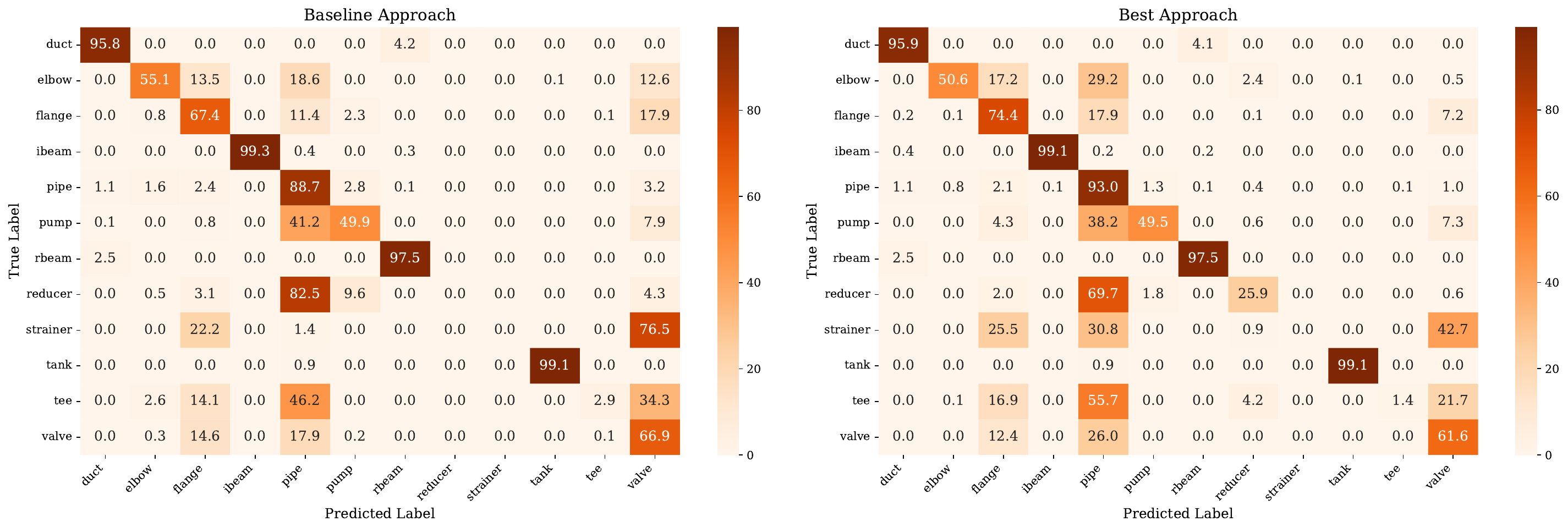}
    \caption{Row-normalized confusion matrices (\%) on the Area~6+12 evaluation split for the cross-entropy baseline (left) and Boundary-CB (right), from an earlier complete 20-vote evaluation pass whose pooled mIoU agrees within 0.1 percentage points with the pass reported in \autoref{tab:per_area_results}. In the baseline, Pipe absorbs a large share of the tail classes with cylindrical primitives: 82.5\% of Reducer points, 46.2\% of Tee points, and 18.6\% of Elbow points are predicted as Pipe. With Boundary-CB: (1) Reducer recall rises from 0\% to 25.9\%, and the share of Reducer points predicted as Pipe falls from 82.5\% to 69.7\%; (2) Pipe recall rises from 88.7\% to 93.0\%, with fewer Pipe points predicted as Elbow (1.6\%$\rightarrow$0.8\%), Flange (2.4\%$\rightarrow$2.1\%), Pump (2.8\%$\rightarrow$1.3\%), and Valve (3.2\%$\rightarrow$1.0\%); (3) Valve IoU rises from 38.5\% to 45.9\% in this pass (38.06\% to 47.32\% in \autoref{tab:per_class_iou}) although Valve recall falls from 66.9\% to 61.6\% and the share of Valve points predicted as Pipe rises from 17.9\% to 26.0\%, so the Valve gain comes mainly from fewer false-positive Valve predictions on Pipe, Elbow, Flange, and Tee points. Tee remains largely confused with Pipe (46.2\%$\rightarrow$55.7\%).}
    \label{fig:confusion_matrix}
\end{figure}

Confusion matrix analysis. The confusion matrices in \autoref{fig:confusion_matrix} reveal three consistent patterns of how Boundary-CB resolves geometric ambiguity through spatial context constraints.

High-confidence classes remain stable. Head and common classes such as \texttt{Duct}, \texttt{RectangularBeam}, \texttt{Ibeam}, and \texttt{Tank} are already well separated in feature space, so Boundary-CB preserves their strong performance while focusing additional learning capacity on ambiguous regions.

Ambiguity-dominated classes improve most. The largest gains appear exactly where primitive-sharing ambiguity is strongest. Reducer improves from 0\% to 21.12\% IoU by reducing confusion with \texttt{Pipe}. Valve improves from 38.06\% to 47.32\%, indicating that neighborhood disagreement is informative at multi-part boundaries. Flange, Pipe, Pump, and Elbow also improve, showing that the entropy signal helps the network separate locally similar cylindrical or assembly-like regions.

Residual failure mode. \texttt{Strainer} remains at 0\% IoU under all losses, indicating an extreme few-shot regime where re-weighting alone is insufficient. \texttt{Tee} improves only marginally, which suggests that rare branching geometries remain difficult even when boundary emphasis is added.

Mechanism interpretation. Three observations are consistent with Boundary-CB's behavior. First, boundary-enhanced learning via entropy focus: the constraint upweights high-entropy regions where neighborhood predictions disagree, which occur at ambiguous transitions such as pipe-reducer and pipe-valve interfaces. Second, more precise tail predictions: in \autoref{fig:confusion_matrix}, the gains come mainly from fewer false-positive tail predictions (e.g., for \texttt{Valve}), whereas several tail classes lose some recall to \texttt{Pipe}; \texttt{Reducer}, whose recall rises from 0\% to 25.9\%, is the exception. Third, scale-sensitive context: although $k=64$ is the best tested global setting, the classwise sweep (one run per setting) suggests that smaller neighborhoods can better preserve locality for compact or branching components such as \texttt{Elbow} and \texttt{Tee}.

\subsection{Qualitative Analysis}
\label{sec:qualitative_analysis}
\begin{figure*}[!htbp]
    \centering
    \includegraphics[width=0.98\linewidth,height=0.74\textheight,keepaspectratio]{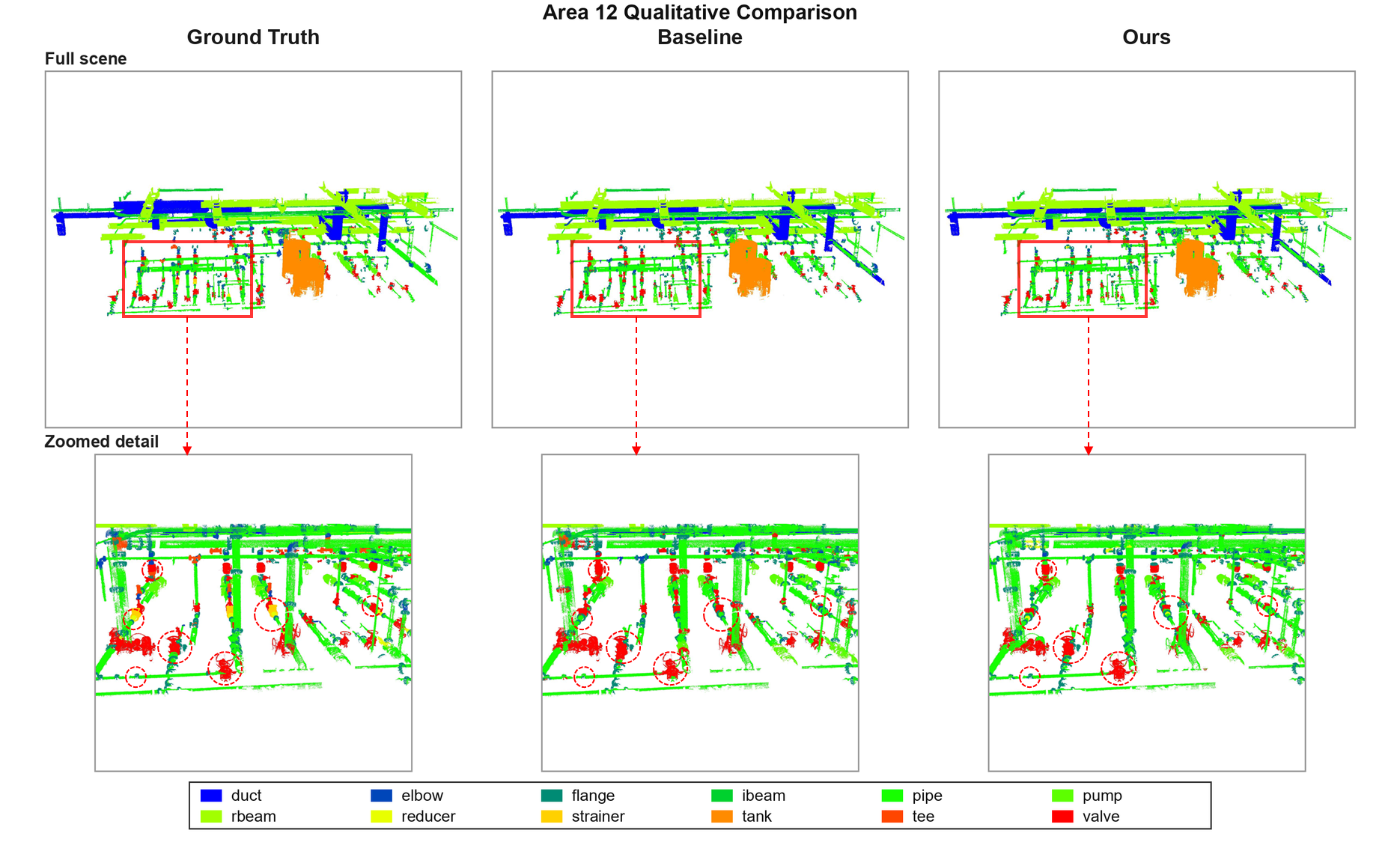}
    \caption{Qualitative segmentation results on Industrial3D Area 12 with paired full-scene and zoomed detail panels. Columns compare ground truth, baseline predictions, and predictions with spatial context constraints. The zoomed detail row focuses on reducer- and valve-rich piping regions where rare functional components share cylindrical or flange-like primitives with dominant pipe structures. Red dashed circles mark representative tail-component regions where the baseline tends to merge or miss compact fittings and our method better preserves or separates them. These circles are visual inspection aids because the improved tail components are statistically rare and spatially small; the quantitative mIoU, per-class IoU, and ablation results provide the primary evidence of improvement. The legend gives the semantic class colors used in all panels.}
    \label{fig:qualitative_results}
\end{figure*}

\autoref{fig:qualitative_results} compares our method with spatial context constraints against the baseline on Area 12, the evaluation area selected for qualitative inspection because it contains reducer- and valve-dense MEP regions. The panel pairs full-scene views with manually selected zoomed details for ground truth, baseline, and our prediction; red dashed circles identify representative rare-component regions for inspection, and the legend maps colors to semantic classes.

Area 12 highlights the practical form of primitive-sharing ambiguity in TLS-based industrial MEP scenes. The baseline tends to absorb reducers, valves, and compact fittings into dominant pipe-like structures because these components share local cylindrical geometry and occur sparsely. The zoomed comparison and red dashed circles make these errors easier to inspect than the full-scene view alone: spatial context constraints recover more of the rare functional components and produce cleaner separation around dense pipe assemblies, although the quantitative tables remain the more reliable basis for judging rare tail-component gains.

These qualitative results demonstrate practical value for MEP Digital Twin applications, supporting more reliable segmentation of safety-critical components for predictive maintenance and facility management.

\subsection{Ablation Studies}
\label{sec:ablation_studies}
\begin{figure}[!htbp]
    \centering
    \includegraphics[width=\linewidth]{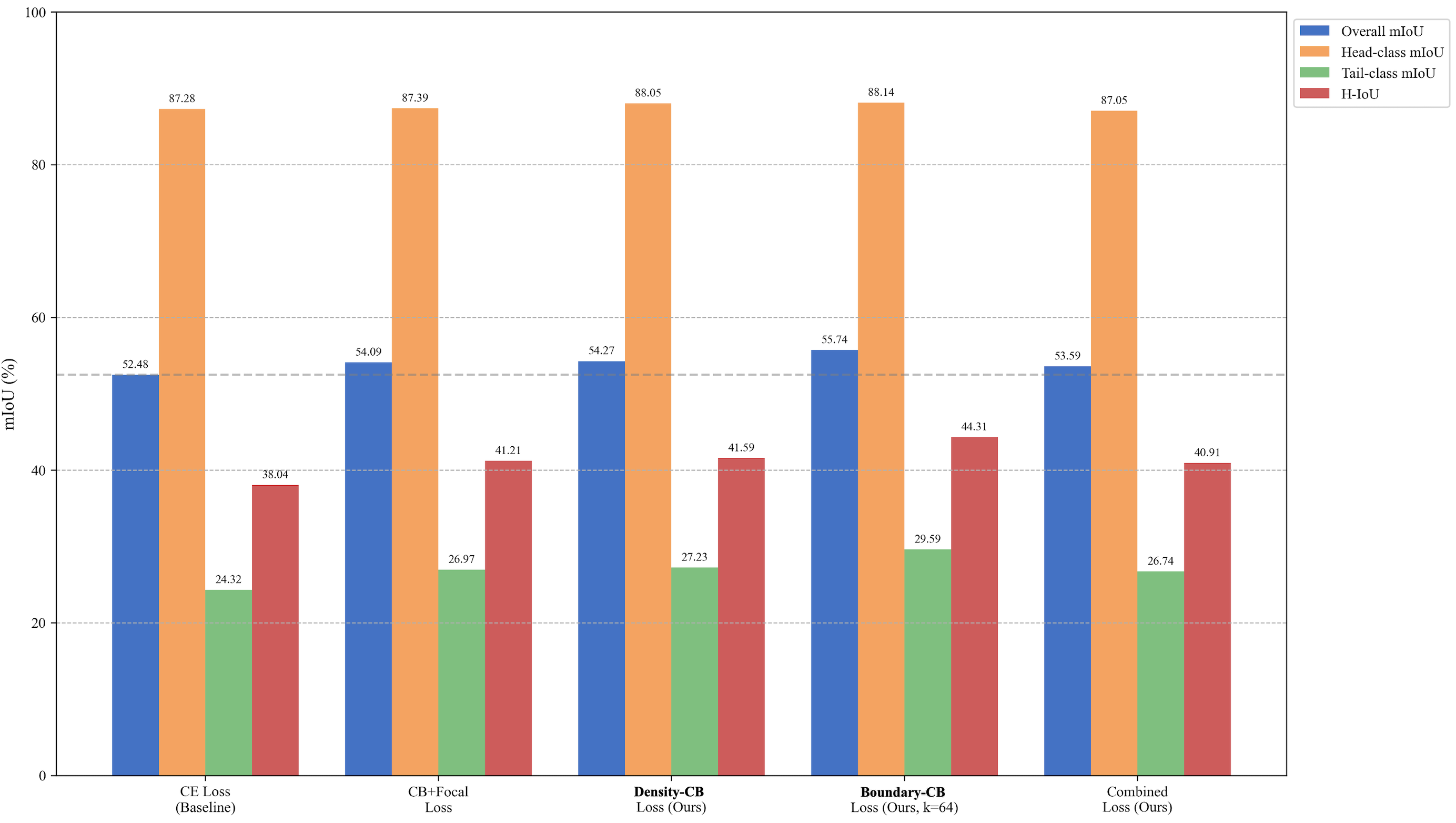}
    \caption{Comparison of the loss variants on the evaluation split: overall mIoU (blue bars), head-class mIoU (orange bars), tail-class mIoU (green bars), and H-IoU (red bars). Methods: (1) CE Baseline: standard cross-entropy achieves 52.48\% overall, 24.32\% tail; (2) +CB+Focal: frequency-based re-weighting improves overall to 54.09\% (+1.61 pp) but tail gains only +2.65 pp (26.97\%), showing statistical correction alone insufficient for geometric ambiguity; (3) +Density-CB: density normalization adds marginal gains (+0.18 pp overall, +0.26 pp tail); (4) +Boundary-CB ($k=64$): spatial context constraint provides the largest tail gain (+5.27 pp over baseline, reaching 29.59\%), indicating that neighborhood prediction consistency helps with primitive-sharing ambiguity. Key finding: Boundary-CB gives the largest tail-class gain on this split. All values are the training-time best-epoch scores of \autoref{tab:main_results}; in the complete evaluation pass of \autoref{tab:per_area_results}, CB+Focal does not exceed the baseline.}
    \label{fig:component_ablation}
\end{figure}

We validate key design choices through systematic ablation studies. \autoref{fig:component_ablation} summarizes component-wise contributions.

\subsubsection{Impact of Spatial Context Parameters}
\autoref{tab:ablation_bc_k} shows Boundary-CB peaks at $k=64$ for overall and tail mIoU, making this spatial context window the best tested global operating point on Industrial3D. The verified classwise sweep is more heterogeneous: \texttt{Reducer}, \texttt{Valve}, and \texttt{Pump} favor $k=64$, \texttt{Elbow} peaks at $k=32$, \texttt{Flange} peaks at $k=16$, and \texttt{Tee} reaches its best Boundary-CB score at $k=8$. This shows that the relevant factor is the spatial extent of the ambiguous region rather than a single class taxonomy label. For Density-CB, \autoref{tab:ablation_radius} shows that $r=0.2$ m, the smallest radius tested, gives the best result; larger radii dilute the density signal by including irrelevant points.

\begin{table}[!htbp]
\centering
\caption{Ablation on the Boundary-CB spatial context window ($k$). The parameter $k$ controls the spatial context radius for measuring neighborhood prediction consistency. Performance peaks at $k=64$, achieving the best tested overall mIoU (55.74\%). This shows that context scale affects geometric disambiguation; for example, \texttt{Reducer} improves from 1.03\% at $k=32$ to 21.12\% at $k=64$. However, excessively large context ($k=128$) introduces noise from irrelevant points and degrades performance. \autoref{sec:discussion} analyzes the trade-offs between different $k$ values for specific component types. The best tested setting is in bold.}
\label{tab:ablation_bc_k}
\begin{tabular}{lcccc}
\toprule
k (neighbors) & mIoU(\%) & mIoU$_{\text{head}}$(\%) & mIoU$_{\text{tail}}$(\%) & H-IoU(\%) \\
\midrule
8 & 53.74 & 86.05 & 27.09 & 41.21 \\
16 & 53.89 & 88.80 & 26.30 & 40.58 \\
32 & 54.00 & 88.08 & 26.53 & 40.78 \\
\textbf{64} & \textbf{55.74} & \textbf{88.14} & \textbf{29.59} & \textbf{44.31} \\
128 & 52.88 & 85.56 & 26.31 & 40.24 \\
\bottomrule
\end{tabular}
\end{table}

\begin{table}[!htbp]
\centering
\caption{Ablation on the radius parameter ($r$) for the Density-CB constraint. The radius $r$ defines the neighborhood for computing local point density normalization. Among the tested radii, $r=0.2$ m gives the best mIoU (54.27\%) and tail-mIoU (27.23\%); smaller radii were not tested. This radius corresponds to the typical size of small MEP components, suggesting that class-appropriate geometric scales are crucial for density normalization as discussed in \autoref{sec:methodology}. Larger radii degrade performance by including irrelevant points. The best tested setting is in bold. $^{\dagger}$This run also used different focal and density-weighting settings, so it is not a controlled radius comparison.}
\label{tab:ablation_radius}
\begin{tabular}{lcccc}
\toprule
Radius r (m) & mIoU(\%) & mIoU$_{\text{head}}$(\%) & mIoU$_{\text{tail}}$(\%) & H-IoU(\%) \\
\midrule
\textbf{0.2} & \textbf{54.27} & \textbf{88.05} & \textbf{27.23} & \textbf{41.59} \\
0.5$^{\dagger}$ & 53.34 & 88.30 & 26.58 & 40.86 \\
0.6 & 52.56 & 88.30 & 24.79 & 38.71 \\
\bottomrule
\end{tabular}
\end{table}

\subsubsection{Constraint Parameter Settings}
\begin{figure}[!htbp]
    \centering
    \includegraphics[width=\linewidth]{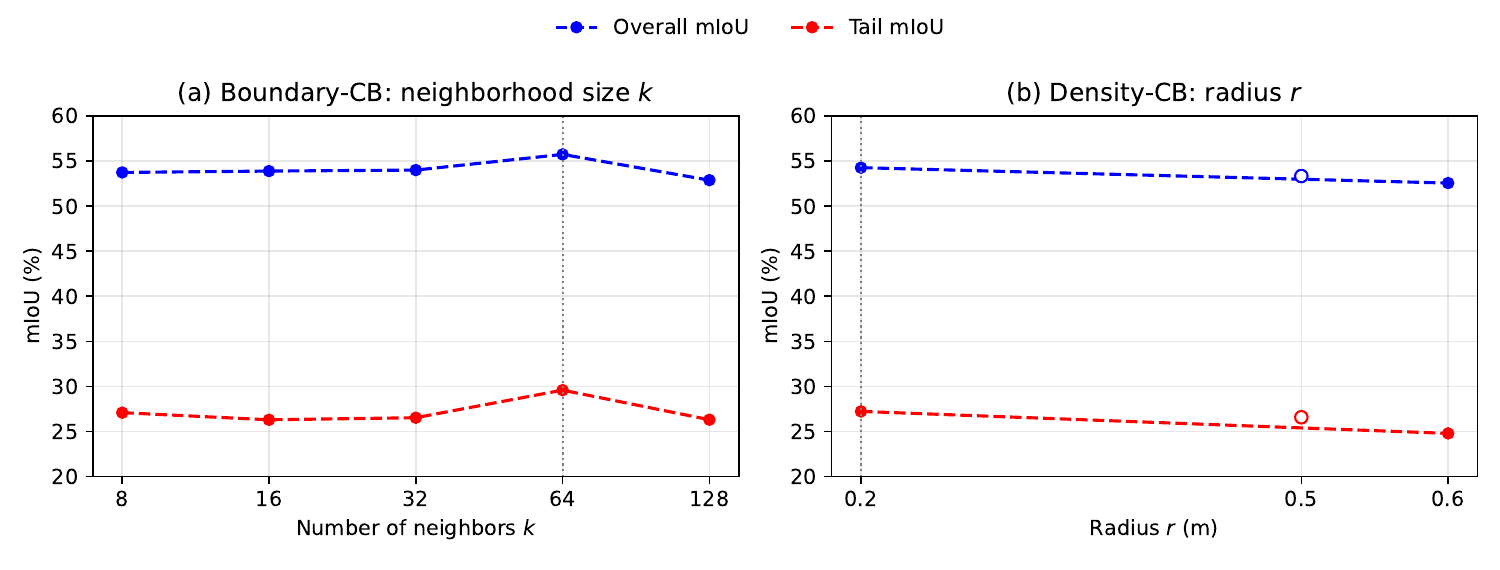}
    \caption{Sensitivity of Boundary-CB to the neighborhood size $k$ (a) and of Density-CB to the density radius $r$ (b) on the evaluation split, showing overall mIoU (blue) and tail-class mIoU (red). Values are those of \autoref{tab:ablation_bc_k} and \autoref{tab:ablation_radius}; dotted lines mark the best tested settings, $k=64$ and $r=0.2$ m. The hollow markers at $r=0.5$ m show the run that also used different focal and weighting settings (\autoref{tab:ablation_radius}), so it is not joined to the controlled points.}
    \label{fig:ablation_curves}
\end{figure}

The CB coefficient and the Boundary-CB modulation strength are fixed at $\beta=0.9999$, a value commonly used for severe imbalance~\cite{cui2019class}, and $\alpha=1.0$; neither was tuned. \autoref{fig:ablation_curves} summarizes the neighborhood-size and radius sweeps. Training overhead is negligible for Density-CB and Boundary-CB. This overhead is small relative to the observed +5.27~pp tail-mIoU gain on the predefined split.

\section{Discussion}
\label{sec:discussion}
This study shows that spatial context constraints improve the treatment of geometric ambiguity in long-tailed TLS-based water-treatment MEP point cloud segmentation. By introducing neighborhood prediction consistency as a constraint mechanism, our approach achieves substantial gains on safety-critical components while maintaining head-class performance on the predefined Industrial3D evaluation split. This addresses one aspect of the dual crisis that hinders practical deployment in water-treatment MEP settings.

Compared to previous long-tailed learning methods, this work contributes on four levels:

Problem formalization. We identify and characterize the ``dual crisis'' in Industrial3D water-treatment MEP data: extreme statistical imbalance (215:1 ratio) combined with geometric ambiguity where six of seven tail classes (86\%) share cylindrical primitives with head classes. We distinguish two mechanisms of geometric ambiguity—\emph{compositional ambiguity} in multi-part assemblies (e.g., \texttt{Valve}, \texttt{Pump}) and \emph{primitive-similarity ambiguity} in pipe-like objects (e.g., \texttt{Reducer}, \texttt{Elbow}). This provides a domain-specific basis for geometry-aware long-tailed 3D learning.

Boundary-CB leverages neighborhood entropy to identify and emphasize ambiguous boundaries, while Density-CB normalizes for scan-dependent density variations. Both constraints extend the CB framework to incorporate spatial geometry. The revised classwise sensitivity analysis shows that $k=64$ is the best global operating point on Industrial3D, but not a universal classwise optimum. Reducer, Valve, and Pump benefit from the wider context, whereas compact or branching components such as Elbow and Tee prefer smaller neighborhoods. Spatial scale therefore has to match the extent of the ambiguous region rather than follow a single rule for all tail classes.

This mechanism-level framing connects the method to engineering knowledge representation. Boundary-CB transforms an MEP topology prior into a training signal by emphasizing uncertain transitions where functional component boundaries coincide with primitive changes, such as pipe-to-reducer and pipe-to-valve interfaces. Density-CB transforms TLS sensor knowledge into a training signal by compensating for scan-dependent density variation that persists after voxel/grid subsampling. Together, the constraints convert domain knowledge about assembly geometry and acquisition physics into principled loss modulation, advancing engineering-knowledge-informed learning rather than generic algorithmic improvement alone.

Unlike specialized network designs, the proposed constraints function as plug-and-play modules that require no architectural changes, although this study evaluates them with ResPointNet++ only. This enables integration into existing TLS-based MEP segmentation pipelines with negligible training overhead and no inference cost.

For engineering applications, the constraints support more reliable identification of functional components such as reducers and valves, which are critical for Digital Twin construction and automated knowledge extraction from facility data. Correctly segmented components could also serve as inputs to point-cloud-centered multimodal large language models, which describe the functions of indoor building components and answer engineering questions about them~\cite{jing2026geometric}. By achieving 21.7\% relative improvement on tail-class performance while preserving head-class accuracy (88.14\%), the method avoids the typical head-tail trade-off that undermines practical utility in long-tailed benchmarks.

Several avenues remain for extending this work. First, the current evidence is scoped to one predefined Industrial3D split, a controlled ResPointNet++ loss ablation, and same-dataset benchmark context across representative backbones. Areas~6 and~12 differ substantially in MEP function, layout, and class composition, but the pooled result is not evidence of cross-dataset generalization, and the use of this split for validation further limits an independent generalization interpretation. A future study should freeze the method before evaluation and test it through grouped cross-area validation, additional MEP datasets, and additional backbones under comparable protocols. The per-area evaluation suggests that the benefit depends on the evaluated area: Boundary-CB gives its largest gains in Area~12, where primitive-sharing tail components are common, and scores slightly below the other models in Area~6, where such components are rare and both rooms come from the same facility as five training areas (\autoref{tab:per_area_results}); checkpoint selection also relied mainly on Area~12. With two evaluation areas these explanations cannot be separated, and gains may be small in areas dominated by structural members. Each method was also trained once, so the Area~6 differences, all within 1.5 percentage points, cannot be separated from training variability, and the Area~12 gain likewise rests on a single training run per method. Ultra-few-shot categories (e.g., \texttt{Strainer}, 0\% IoU) require additional techniques such as explicit structural priors (directional continuity for pipes, topological signatures for branching) or few-shot meta-learning. The classwise k-sweep also suggests, from one run per setting, that one neighborhood size does not suit every component equally: Reducer, Valve, and Pump benefit from $k=64$, Elbow peaks at $k=32$, Tee peaks at $k=8$, and Flange peaks at $k=16$ within the verified Boundary-CB sweep. This heterogeneity motivates \emph{scale-adaptive context} mechanisms that adjust the spatial window size based on component geometry and sampling density. Extending spatial context constraints to temporal consistency (multi-scan temporal coherence) and cross-modal learning (fusing RGB, thermal, or LiDAR intensity channels) could further improve robustness in complex MEP environments.

\section{Conclusion}
\label{sec:conclusion}
This study addressed geometric ambiguity in long-tailed TLS-based water-treatment MEP point cloud segmentation, a critical bottleneck for automated knowledge extraction and Digital Twin construction. The Industrial3D analysis shows why frequency-only methods are insufficient in this setting: head classes account for 77\% of points and are dominated by long, continuous structures such as pipes, beams, and ducts, whereas tail classes include multi-part assemblies such as \texttt{Valve} and \texttt{Pump} as well as pipe-like components such as \texttt{Reducer} and \texttt{Elbow}. To address this dual crisis of statistical imbalance and primitive-sharing ambiguity, we extended the Class-Balanced framework with spatial context constraints based on neighborhood prediction consistency and density normalization.

On the predefined Industrial3D Area~6+12 evaluation split, Boundary-CB achieves 55.74\% mIoU and a 21.7\% relative improvement on safety-critical tail classes (29.59\% vs.\ 24.32\% baseline) while maintaining head-class accuracy (88.14\%). It improves previously difficult components, including \texttt{Reducer} (21.12\% vs.\ 0\% baseline) and \texttt{Valve} (+24.3\% relative), and raises the head--tail harmonic mean IoU from 38.04\% to 44.31\% without the typical head-tail trade-off on that split. Evaluated area by area, the gain comes from Area~12; in Area~6, Boundary-CB is 0.6 percentage points below the baseline in mIoU. Both Boundary-CB and Density-CB integrate into existing point cloud pipelines without network modifications, with negligible training overhead and no inference cost.

This work shows that spatial context constraints can reduce primitive-sharing errors in long-tailed TLS-based MEP point cloud segmentation on Industrial3D, supporting more reliable knowledge extraction and Digital Twin construction in such settings.

\stopmainlinenumbers

\section*{Supplementary Material}
Supplementary material associated with this article (complementary dataset analysis, combined loss function experiments, fixed CB and Boundary-CB parameters, classwise Boundary-CB sensitivity analysis, practical neighborhood-size guidance, area-level class composition with classwise per-area results, and complete Industrial3D dataset statistics) can be found in the supplementary material.

\section*{Acknowledgments}
This research was jointly supported by the China Postdoctoral Science Foundation (grant number 2023M740761), the Natural Science Foundation of Hunan Province, China (grant number 2023JJ40098), the National Natural Science Foundation of China (No. 42471432, 42301536 and 42271479), and the Science and Technology Program of Guangdong (No. 2024B1212080002).

\section*{Data Availability Statement}
 The Industrial3D dataset will be publicly released upon acceptance of the companion benchmark paper currently under review; preview materials, benchmark documentation, and release information are currently available at \url{https://github.com/pointcloudyc/Industrial3D.git}. The source code and trained models for Boundary-CB and Density-CB will be released upon acceptance of the present manuscript at \url{https://github.com/pointcloudyc/LongTail3D.git}. Researchers interested in early access to the dataset or code during the review process are welcome to contact the corresponding author.

\section*{Declaration of Competing Interests}
The authors declare that they have no known competing financial interests or personal relationships that could have appeared to influence the work reported in this paper.

\bibliographystyle{elsarticle-num}
\bibliography{references}
\end{document}